\documentclass{article}

\usepackage[final,nonatbib]{neurips_2024}

\usepackage[utf8]{inputenc} 
\usepackage[T1]{fontenc}    
\usepackage[hidelinks]{hyperref}       
\usepackage{url}            
\usepackage{booktabs}       
\usepackage{amsmath}
\usepackage{amsfonts}       
\usepackage{nicefrac}       
\usepackage{microtype}      
\usepackage{xcolor}         
\usepackage{multicol}
\usepackage{multirow}
\usepackage{graphicx}

\usepackage{changepage}
\usepackage{enumerate}
\usepackage[font=small]{caption}

\usepackage{titletoc}

\newcommand\DoToC{%
  \startcontents
  \printcontents{}{1}{\textbf{Table of Contents}\vskip3pt\hrule\vskip5pt}
  \vskip3pt\hrule\vskip5pt
}
\usepackage{wrapfig}

\newcommand\blfootnote[1]{%
  \begingroup
  \renewcommand\thefootnote{}\footnote{#1}%
  \addtocounter{footnote}{-1}%
  \endgroup
}

\newcommand{\rach}[1]{\textcolor{blue}{#1}}
\def \RR {{\mathbb{R}}}

\def\vvarphi{{\bm{\varphi}}}
\def\vphi{{\bm{\phi}}}
\def \vgamma {{\bm \gamma}}
\newtheorem{remark}{Remark}
\newtheorem{theorem}{Theorem}
\newtheorem{definition}{Definition}

\PassOptionsToPackage{numbers, compress}{natbib}
\usepackage{amsmath}
\usepackage{microtype}
\usepackage{graphicx}
\usepackage{subfigure}
\usepackage{hyperref}
\hypersetup{
    colorlinks=true,
    linkcolor=blue,
    filecolor=magenta,      
    urlcolor=cyan,
    citecolor=black
}
\usepackage{booktabs} 
\usepackage{amsfonts}
\usepackage{bm}
\usepackage{threeparttable}
\usepackage{tablefootnote}

\usepackage{lipsum}
\usepackage{comment}

\usepackage{multirow}
\usepackage{amsmath}

\usepackage{hyperref}

\usepackage{enumitem}

\usepackage{mathtools}

\usepackage{wrapfig}

\usepackage{listings}

\usepackage[english]{babel}

\usepackage{listings}
\usepackage{color}

\definecolor{mygreen}{rgb}{0,0.6,0}
\definecolor{mygray}{rgb}{0.5,0.5,0.5}
\definecolor{mymauve}{rgb}{0.58,0,0.82}

\usepackage{marginnote}
\usepackage{soul} 

\newcommand{\ignore}[1]{}



\usepackage{amsmath,amsfonts,bm}

\def\eqref#1{equation~\ref{#1}}

\def\1{\bm{1}}

\def\va{{\bm{a}}}

\def\vh{{\bm{h}}}

\def\vj{{\bm{j}}}
\def\vk{{\bm{k}}}

\def\vq{{\bm{q}}}

\def\vu{{\bm{u}}}
\def\vv{{\bm{v}}}

\def\vx{{\bm{x}}}
\def\vy{{\bm{y}}}

\def\mA{{\bm{A}}}

\def\mC{{\bm{C}}}

\def\mG{{\bm{G}}}
\def\mH{{\bm{H}}}
\def\mI{{\bm{I}}}

\def\mK{{\bm{K}}}
\def\mL{{\bm{L}}}
\def\mM{{\bm{M}}}

\def\mQ{{\bm{Q}}}

\def\mS{{\bm{S}}}

\def\mU{{\bm{U}}}
\def\mV{{\bm{V}}}
\def\mW{{\bm{W}}}
\def\mX{{\bm{X}}}
\def\mY{{\bm{Y}}}

\DeclareMathAlphabet{\mathsfit}{\encodingdefault}{\sfdefault}{m}{sl}
\SetMathAlphabet{\mathsfit}{bold}{\encodingdefault}{\sfdefault}{bx}{n}
\usepackage{algorithmic}
\usepackage{algorithm}

\title{Unveiling the Hidden Structure of Self-Attention \\via Kernel Principal Component Analysis}

\author{%
  Rachel S.Y. Teo \\
  Department of Mathematics\\
  National University of Singapore\\
  \texttt{rachel.tsy@u.nus.edu} \\
  \And
  Tan M. Nguyen \\
  Department of Mathematics\\
  National University of Singapore\\
  \texttt{tanmn@nus.edu.sg} \\
}

\begin{document}

\maketitle

\begin{abstract}
The remarkable success of transformers in sequence modeling tasks, spanning various applications in natural language processing and computer vision, is attributed to the critical role of self-attention. Similar to the development of most deep learning models, the construction of these attention mechanisms relies on heuristics and experience. In our work, we derive self-attention from kernel principal component analysis (kernel PCA) and show that self-attention projects its query vectors onto the principal component axes of its key matrix in a feature space. We then formulate the exact formula for the value matrix in self-attention, theoretically and empirically demonstrating that this value matrix captures the eigenvectors of the Gram matrix of the key vectors in self-attention. Leveraging our kernel PCA framework, we propose Attention with Robust Principal Components (RPC-Attention), a novel class of robust attention that is resilient to data contamination. We empirically demonstrate the advantages of RPC-Attention over softmax attention on the ImageNet-1K object classification, WikiText-103 language modeling, and ADE20K image segmentation task. The code is publicly available at \texttt{\url{https://github.com/rachtsy/KPCA_code}}.
\end{abstract}

\section{Introduction}

Transformers~\cite{vaswani2017attention} have emerged as the preeminent model for tackling a myriad of challenging problems in natural language processing~\cite{baevski2018adaptive,devlin-etal-2019-bert,al2019character,JMLR:v21:20-074,NEURIPS2020_1457c0d6,NEURIPS2022_8bb0d291,JMLR:v24:22-1144}, computer vision~\cite{dosovitskiy2021an,DBLP:journals/corr/abs-2012-12877,liu2021swin,radford2021learning,khan2022transformers,gal2023an}, reinforcement learning~\cite{chen2021decision,janner2021offline,zheng2022online,lee2022multi}, and other applications~\cite{zhang2019deep,rives2021biological,jumper2021highly,wang2022transtab}. The effectiveness of transformers is rooted in their ability to learn from unlabeled data and take advantage of pre-trained models for downstream tasks that involve diverse data modalities with limited supervision~\cite{radford2018improving,radford2019language,NEURIPS2019_dc6a7e65,zhuang-etal-2021-robustly,pmlr-v139-rao21a}. At the core of the transformer's success lies the self-attention mechanism, which serves as the fundamental building block of a transformer model. This mechanism enables each token in a sequence to aggregate information from other tokens by computing a weighted average based on similarity scores between their feature representations. Facilitating dynamic interactions among tokens, this attention mechanism allows tokens to selectively attend to others, thereby attaining a contextual representation~\cite{bahdanau2014neural,nguyen2023probabilistic,parikh-etal-2016-decomposable,lin2017a}. The flexibility in capturing diverse syntactic and semantic relationships is an important factor contributing to the success of transformers~\cite{tenney-etal-2019-bert,voita-etal-2019-analyzing,hewitt-liang-2019-designing,vig-belinkov-2019-analyzing,nguyen2022head}.\blfootnote{Please correspond to: \texttt{tanmn@nus.edu.sg}}

{\bf Self-Attention.} For a given input sequence $\mX:=[\vx_1,\cdots,\vx_N]^\top\in \RR^{N\times D_x}$ of $N$ feature vectors, self-attention transforms $\mX$ into the output sequence $\mH$ in the following two steps:

{\it \underline{Step 1}:} The input sequence $\mX$ is projected into the query matrix $\mQ$, the key matrix $\mK$, and the value matrix $\mV$ via three linear transformations
\begin{align}
\mQ=\mX\mW_Q^\top; \mK=\mX\mW_K^\top; \mV=\mX\mW_V^\top, 
\end{align}
where $\mW_Q,\mW_K\in \RR^{D\times D_x}$, and $\mW_V\in \RR^{D_v\times D_x}$ are the weight matrices. We denote $\mQ:=[\vq_1,\cdots,\vq_N]^\top, \mK:=[\vk_1,\cdots,\vk_N]^\top$, and $\mV:=[\vv_1,\cdots,\vv_N]^\top$, where the vectors $\vq_i,\vk_i,\vv_i$ for $i=1,\cdots,N$ are the query, key, and value vectors, respectively.
    
{\it \underline{Step 2}:} The output sequence $\mH:=[\vh_1,\cdots,\vh_N]^\top$ is then computed as follows
\begin{equation}\label{eq:attention-mat}
\mH={\rm softmax}\Big({\mQ}{\mK}^\top /{\sqrt{D}}\Big)\mV :=\mA{\mV},
\end{equation}
where the softmax function is applied to each row of the matrix $\mQ\mK^\top/\sqrt{D}$. The matrix $\mA:={\rm softmax}\Big(\frac{{\mQ}{\mK}^\top }{\sqrt{D}}\Big) \in \RR^{N\times N}$ and its component $a_{ij}$ for $i,\,j=1,\cdots,N$  are called the attention matrix and attention scores, respectively. For each query vector $\vq_i$ for $i=1,\cdots,N$, an equivalent form of Eqn.~(\ref{eq:attention-mat}) to compute the output vector $\vh_i$ is given by 
\begin{equation}\label{eq:attention-vec}
\vh_i=\sum_{j=1}^N{\rm softmax}\Big({\vq}_i^\top{\vk}_j/\sqrt{D}\Big){\vv}_j.
\end{equation}
The self-attention computed by Eqn.~(\ref{eq:attention-mat}) and~(\ref{eq:attention-vec}) is called the scaled dot-product or softmax attention. In our paper, we call a transformer that uses this attention the softmax transformer. The structure that the attention matrix $\mA$ learns from training determines the ability of the self-attention to capture contextual representations for each token. 
Despite their impressive achievements, the development of most attention layers rely on intuitions and heuristic approaches. The quest for a systematic and principled framework for studying and synthesizing attention layers has remained challenging.

{\bf Contribution.} We study and analyze self-attention in transformers from the perspective of kernel principal component analysis (kernel PCA). In particular, we develop a novel connection between self-attention and kernel PCA, showing that \emph{self-attention projects its query vectors $\vq_i$, $i=1,\dots,N$, onto principal component axes of the key matrix $\mK$ in a feature space}.  We then inspect the structure of the value matrix $\mV$ of self-attention suggested by our kernel PCA framework, validating \emph{$\mV$ captures the eigenvectors of the Gram matrix of the key vectors $\vk_j$,$j=1,\dots,N$}. Using our framework, we then propose a new class of robust attention, namely the Attention with Robust Principal Components (RPC-Attention). Our contribution is three-fold.
\begin{enumerate}[leftmargin=25pt]
    \item We derive self-attention from kernel PCA, showing that the attention outputs are projections of the query vectors onto the principal components axes of the key matrix $\mK$ in a feature space. 
    \item We discover and validate that the value matrix $\mV$ in self-attention captures the eigenvectors of the Gram matrix of the key vectors $\vk_j$,$j=1,\dots,N$.
    \item We develop the Attention with Robust Principal Components (RPC-Attention), a new attention mechanism that is resilient to data contamination, using our kernel PCA framework.
\end{enumerate}
We empirically demonstrate the benefits of RPC-Attention on the ImageNet-1K object classification, ADE20K image segmentation, and large scale WikiText-103 language modeling tasks. We further illustrate the robustness of RPC-Attention through our evaluations on popular, standard robustness benchmarks, as well as various white and black box adversarial attacks on ImageNet-1K images, 15 different types of corruptions on the ADE20K dataset, and word swap attack on WikiText-103. 

\section{Principal Component Analysis of Attention}
\label{sec:kpca-attn}
In this section, we will derive attention from kernel PCA. Suppose we are given a dataset $M = \{\vk_1,\dots,\vk_N\}\subset \RR^{D}$. Here, $\vk_1,\dots,\vk_N$ are attention keys in self-attention. As in kernel PCA, we first project these data points into a feature space using a feature map $\vvarphi(\vx):= \vphi(\vx)/g(\vx)$, where $\vphi(\cdot)$ is a nonlinear transformation from $\RR^{D}$ to $\RR^{D'}$, and $g(\cdot)$ is a vector-scalar function that computes a scaling factor for $\vphi(\vx)$. We center the projected data as follows:
\begin{align}
    \tilde{\vvarphi}(\vk_j) =  \vvarphi(\vk_j) - \frac{1}{N}\sum_{j'=1}^{N}\vvarphi(\vk_{j'}).
\end{align}
Letting $\mC$ be the covariance matrix of the centered data in feature space, its eigenvector expansion is 
\begin{align}
\mC = \frac{1}{N}\sum_{j=1}^{N}\tilde{\vvarphi}(\vk_j)\tilde{\vvarphi}(\vk_j)^{\top}; \quad
\mC\vu_d = \lambda_d \vu_d, \,\,\, d=1,\dots,D_v.
\label{eqn:eigen_exp}
\end{align} 
Plugging $\mC$ into~(\ref{eqn:eigen_exp}), we obtain
\begin{align}
    \frac{1}{N}\sum_{j=1}^{N}\tilde{\vvarphi}(\vk_j)\{\tilde{\vvarphi}(\vk_j)^{\top}\vu_d\} = \lambda_d \vu_d.
\label{eqn:eigen_exp_2}
\end{align}
Thus, provided that $\lambda_d > 0$, the eigenvector $\vu_d$ is given by a linear combination of the $\tilde{\vvarphi}(\vk_j)$ and with $a_{dj} = \frac{1}{N\lambda_d}\tilde{\vvarphi}(\vk_j)^{\top}\vu_d$, can be written as
\begin{align}
\vu_d &= \sum_{j=1}^N a_{dj}\tilde{\vvarphi}(\vk_j).
\label{eqn:linear_exp}
\end{align}

\subsection{Deriving Attention from Kernel PCA}
\label{sec:deriving-attn}
In order to derive self-attention from kernel PCA, we consider the query vector $\vq_i$ in self-attention as a new data point. The projection $\vh_i$ of a new test point $\vq_i$ onto the principal components $\vu_d$ in Eqn.~(\ref{eqn:linear_exp}), for $d=1,\dots,D_v$, is given by
\begin{align}
\label{eqn:derive_h}
\vh_i(d) &= \vvarphi(\vq_i)^{\top} \vu_d=\sum_{j=1}^N a_{dj}\vvarphi(\vq_i)^{\top}\tilde{\vvarphi}(\vk_j) =\sum_{j=1}^N a_{dj}\vvarphi(\vq_i)^{\top}(\vvarphi(\vk_j) - \frac{1}{N}\sum_{j'=1}^{N}\vvarphi(\vk_{j'}))\nonumber \\ 
&=\sum_{j=1}^N \frac{k(\vq_i, \vk_j)}{g(\vq_i)}\frac{a_{dj}}{g(\vk_j)} - \frac{1}{N}\sum_{j'=1}^{N}\frac{k(\vq_i, \vk_{j'})}{g(\vq_i)} \sum_{j=1}^N\frac{a_{dj}}{g(\vk_{j'})}\nonumber \\
&=\sum_{j=1}^N \frac{k(\vq_i, \vk_j)}{g(\vq_i)}\Big(\frac{a_{dj}}{g(\vk_j)} - \frac{1}{N}\sum_{j'=1}^N\frac{a_{dj'}}{g(\vk_{j})}\Big) =\sum_{j=1}^N \frac{k(\vq_i, \vk_j)}{g(\vq_i)}v_{dj}, 
\end{align}
where the kernel $k(\vx, \vy) := \vphi(\vx)^{\top}\vphi(\vy)$ and $v_{dj}:=\frac{a_{dj}}{g(\vk_j)} - \frac{1}{N}\sum_{j'=1}^N\frac{a_{dj'}}{g(\vk_{j})}$. We further let the self-attention's value vectors $\vv_j=[v_{1j}, \cdots, v_{D_{v}j}] \in \RR^{D_{v}\times1}$, $j=1,\hdots,N$, and rewrite the projection $\vh_i$ as $\vh_i =\sum_{j=1}^N {k(\vq_i, \vk_j)}/{g(\vq_i)}\vv_j$.
Selecting $g(\vx) := \sum_{j=1}^{N} k(\vx,\vk_j)$ and $k(\vx, \vy) = \exp(\vx^{\top}\vy/\sqrt{D})$, we obtain a formula of an attention:
\begin{align}
\label{eqn:pca-attn-2}
\vh_i =\sum_{j=1}^N \frac{k(\vq_i, \vk_j)}{\sum_{j'=1}^{N} k(\vq_i,\vk_{j'})}\vv_j=\sum_{j=1}^N{\rm softmax}\Big({\vq}_i^\top{\vk}_j/\sqrt{D}\Big)\vv_j.  
\end{align}

{\bf Recovering Self-Attention:} Eqn.~(\ref{eqn:pca-attn-2}) matches the exact formula of a self-attention as in Eqn.~(\ref{eq:attention-vec}). Thus, we can view outputs $\vh_{i}$ of self-attention as projections of the query vectors $\vq_{i}$, $i=1,\dots,N$, onto $D_v$ principal components in a feature space: 
\begin{align}
\label{eqn:kpca_proj}
\mH = [\vvarphi(\vq_1),\dots,\vvarphi(\vq_N)]^{\top}[\vu_1,\dots,\vu_{D_v}].
\end{align}


{\bf Computing the Value Vectors:}
As derived above, the self-attention's value vectors $\vv_j$ are given by: $\vv_j=[v_{1j}, \cdots, v_{D_{v}j}] \in \RR^{D_{v}\times1}$, $j=1,\hdots,N$, where $v_{dj}:=\frac{a_{dj}}{g(\vk_j)} - \frac{1}{N}\sum_{j'=1}^N\frac{a_{dj'}}{g(\vk_{j})}$, $d=1,\dots,D_v$. Since $g(\vk_j)$ can be calculated as $g(\vk_j) = \sum_{j'=1}^{N} k(\vk_j,\vk_{j'})$, in order to compute $\vv_j$, we need to determine the coefficients $a_{1j}, \dots, a_{D_vj}$ for $j=1,\hdots,N$.

We define $\tilde{k}_{\vvarphi}(\vx, \vy) := \tilde{\vvarphi}(\vx)^{\top}\tilde{\vvarphi}(\vy)$ and the Gram matrix $\widetilde{\mK}_{\vvarphi} \in \RR^{N \times N}$ with elements $\widetilde{\mK}_{\vvarphi}(i,j) = \tilde{k}_{\vvarphi}(\vk_i, \vk_j)$. Substituting the linear expansion in Eqn.~(\ref{eqn:linear_exp}) into~(\ref{eqn:eigen_exp_2}), we attain
\begin{align}
\frac{1}{N}\sum_{j=1}^{N}\tilde{\vvarphi}(\vk_j)\tilde{\vvarphi}(\vk_j)^{\top}\sum_{j'=1}^N a_{dj'}\tilde{\vvarphi}(\vk_{j'}) = \lambda_d \sum_{j=1}^N a_{dj}\tilde{\vvarphi}(\vk_j). \nonumber
\end{align}
 We multiply both sides of the above by $\tilde{\vvarphi}(\vk_
\ell)^{\top}$ to obtain $\widetilde{\mK}_{\vvarphi}^{2}\va_d = \lambda_d N \widetilde{\mK}_{\vvarphi} \va_d$,
with the column vector $\va_d = [a_{d1},\cdots,a_{dN}]^\top\in\RR^{N \times 1}$. We compute $\va_d$ by solving 
\begin{align}
\label{eqn:eig_K}
\widetilde{\mK}_{\vvarphi}\va_d = \lambda_d N \va_d.
\end{align}
The calculation of  $\widetilde{\mK}_{\vvarphi}$ is provided in Remark~\ref{rm:comp_gram}. We summarize our results in the following theorem.
\begin{theorem}[Softmax Attention as Principal Component Projections]
\label{theorem:attenion-pca}
Given a set $M$ of key vectors, $M := \{\vk_1,\dots,\vk_N\}\subset \RR^{D}$, a kernel $k(\vx, \vy) := \exp(\vx^{\top}\vy/\sqrt{D})$, and a vector-scalar function $g(\vx) := \sum_{j=1}^{N} k(\vx,\vk_j)$, self-attention performs kernel PCA and projects a query vector $\vq_i \in \RR^{D}$ onto principal component axes of $M$ in an infinite dimensional feature space $\vvarphi$ as follows
\begin{align}
\vh_i &= \sum_{j=1}^N{\rm softmax}\Big({\vq}_i^\top{\vk}_j/\sqrt{D}\Big)\vv_j. \nonumber
\end{align}
The feature space $\vvarphi$ is induced by the kernel $k_{\vvarphi}(\vx, \vy) := \frac{k(\vx,\vy)}{g(\vx)g(\vy)}$, and the value vectors $\vv_j=[v_{1j}, \dots, v_{D_{v}j}] \in \RR^{D_{v}\times1}$, $j=1,\hdots,N$, where $v_{dj}:=\frac{a_{dj}}{g(\vk_j)} - \frac{1}{N}\sum_{j'=1}^N\frac{a_{dj'}}{g(\vk_{j})}$, $d=1,\dots,D_v$. The column vectors $\va_d = [a_{d1},\dots,a_{dN}]^\top\in\RR^{N \times 1}$ can be determined by solving the eigenvalue problem defined in Eqn.~(\ref{eqn:eig_K}). This constraint on $\vv_j$ can be relaxed by letting the self-attention learn $\vv_j$ from data via a linear projection of the input $\vx_j$, i.e., $\vv_j = \mW_V\vx_j$ where $\mW_V$ is a learnable matrix.
\end{theorem}

\begin{remark}[Calculating the Gram Matrix $\widetilde{\mK}_{\vvarphi}$]
\label{rm:comp_gram}
{\normalfont
In the eigenvalue problem defined in Eqn.~(\ref{eqn:eig_K}), the centered Gram matrix $\widetilde{\mK}_{\vvarphi}$ can be computed from the uncentered Gram matrix ${\mK}_{\vvarphi}$ with elements $\mK_{\vvarphi}(i,j) = k_{\vvarphi}(\vk_i, \vk_j) = \vvarphi(\vk_i)^{\top}\vvarphi(\vk_j)$. In particular, $\widetilde{\mK}_{\vvarphi} = {\mK}_{\vvarphi} - \textbf{1}_{N}{\mK}_{\vvarphi} - {\mK}_{\vvarphi}\textbf{1}_{N} + \textbf{1}_{N}{\mK}_{\vvarphi}\textbf{1}_{N}$, where $\textbf{1}_{N}$ denotes the $N \times N$ matrix in which every element takes the value $1/N$~\cite{bishop2006pattern} (See Appendix~\ref{appx:comp_gram}). Here, the kernel $k_{\vvarphi}(\vx, \vy) := \frac{k(\vx,\vy)}{g(\vx)g(\vy)}$.
}
\end{remark}

\begin{remark}[Determining $D_v$]
\label{rm:D_v}
{\normalfont
The feature space $\vvarphi$ is infinite dimensional, so we can find infinitely many principal components. However, the number of nonzero eigenvalues of $\mC$ in Eqn.~(\ref{eqn:eigen_exp}) cannot exceed $N$, the number of data points, since $\mC$ has rank at most equal to $N$. Notice that only principal components corresponding to nonzero eigenvalues are used for projections in kernel PCA. Thus, $D_v$, the number of principal components used for projections as in Eqn.~(\ref{eqn:kpca_proj}), must be less than or equal to $N$, i.e., $D_v \le N$.
}
\end{remark}

\begin{remark}[Parameterization of the Value Matrix $\mV$]
\label{rm:parameterization}
{\normalfont
Different parameterizations of the value matrix $\mV$ can result in different self-attention architectures. For instance, the projection $\vh_i$ of a query vector $\vq_i$ onto the principal components $\vu_d$, $d=1,\dots,D_v$, in Eqn.~(\ref{eqn:derive_h}) can be rewritten as 
\begin{align}
\vh_i(d) &= \sum_{j=1}^N \frac{k(\vq_i, \vk_j)}{g(\vq_i)}\Big(\frac{a_{dj}}{g(\vk_j)} - \frac{1}{N}\sum_{j'=1}^N\frac{g(\vk_{j'})}{g(\vk_{j})}\frac{a_{dj'}}{g(\vk_{j'})}\Big). \nonumber
\end{align}
Letting $v_{dj}:=\frac{a_{dj}}{g(\vk_j)}$ and $s_{jj'}=\frac{g(\vk_{j'})}{g(\vk_{j})}$, we obtain
\begin{align}
\vh_i(d) &= \sum_{j=1}^N \frac{k(\vq_i, \vk_j)}{g(\vq_i)}\Big(v_{dj} -  \frac{1}{N}\sum_{j'=1}^Ns_{jj'}v_{dj'}\Big). \nonumber
\end{align}
Following the same derivation as above, we can write the projection $\vh_i$ as an attention
\begin{align}
\vh_i &= \sum_{j=1}^N{\rm softmax}\Big({\vq}_i^\top{\vk}_j/\sqrt{D}\Big)\Big(\vv_j - \frac{1}{N}\sum_{j'=1}^Ns_{jj'}\vv_{j'} \Big). \nonumber
\end{align}
The matrix form of this new attention form is as follows:
\begin{align}\label{eqn:pca-attn-reparam}
\mH &={\rm softmax}\Big({\mQ}{\mK}^\top /{\sqrt{D}}\Big)(\mI - \mS)\mV,
\end{align}
where $\mI$ is an identity matrix, and $\mS$ is the matrix whose elements $\mS(j,j') =  \frac{1}{N}s_{jj'}$. 
We name the self-attention defined by Eqn.~(\ref{eqn:pca-attn-reparam}) the Scaled Attention. Even though the softmax attention in (\ref{eq:attention-mat}) and the Scaled Attention in (\ref{eqn:pca-attn-reparam}) are mathematically equivalent according to our kernel PCA framework, the training procedure might cause the self-attention models that are derived from different
parameterizations to have different performances.
}
\end{remark}

\subsection{Analysis on the Convergence of Self-Attention Layers to Kernel PCA}
In this section, we discuss empirical justifications that after training, the value vectors $\vv_j$ parameterized as a 1-layer linear network, i.e., $\vv_j = \mW_V\vx_j$, $j=1,\dots,N$, in self-attention converge to the values predicted by our kernel PCA framework in Theorem~\ref{theorem:attenion-pca}. In other words, we provide evidence that the self-attention layers in transformers try to learn their value vectors   $\vv_j$ to perform the kernel PCA.
\subsubsection{Projection Error Minimization}
\label{subsubsec:J_proj}
PCA can be formulated based on projection error minimization as well. In particular, PCA minimizes the average projection cost defined as the mean squared distance between the original data points and their projections~\cite{pearson1901liii}. Given our kernel PCA framework in Theorem~\ref{theorem:attenion-pca}, this implies that self-attention minimizes the following projection error
\begin{align}
\label{eqn:proj-cost}
J_{\text{proj}} = \frac{1}{N}\sum_{i=1}^{N}\|\vvarphi(\vq_i) - \sum_{d=1}^{D_v}h_{di}\vu_d\|^{2}
= \frac{1}{N}\sum_{i=1}^{N}\left(\|\vvarphi(\vq_i)\|^{2} - \|\vh_i\|^{2}\right). 
\end{align}
\begin{wrapfigure}[17]{r}{.45\linewidth}
\vspace{-0.2in}
\small
\centering
\includegraphics[width=1.0\linewidth]{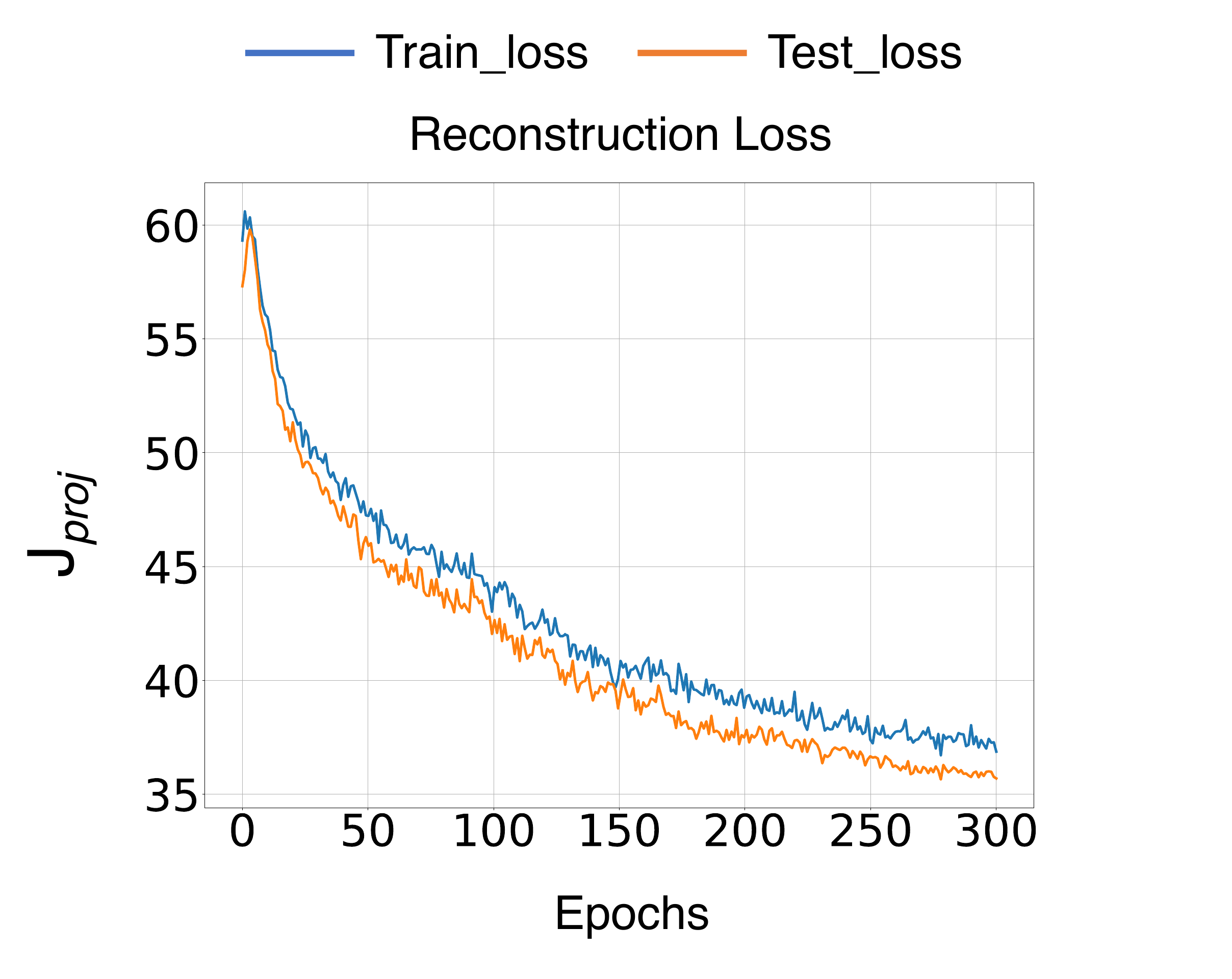}
  \caption{\small Projection loss  vs.\ training epochs of ViT-tiny model. The reconstruction loss is averaged over the batch, heads, and layers. The downward trend suggests that the model is implicitly minimizing this projection loss.}
  \label{fig:proj_cost}
\end{wrapfigure}
In the derivation above, we leverage the
orthonormality of 
$\{\vu_d\}_{d=1}^{D_v}$ and $h_{di} = \vvarphi(\vq_i)^{\top}\vu_{d}$. 
Here, notice that we can compute $\|\vvarphi(\vq_i)\|^{2}$ from $\vq_{i}$ and $\{\vk_{j}\}_{j=1}^{N}$ as $\|\vvarphi(\vq_i)\|^{2} = \exp(\vq_{i}^{\top}\vq_{i}/\sqrt{D})/(\sum_{j=1}^{N}\exp(\vq_{i}^{\top}\vk_{j}/\sqrt{D}))^{2}$. In Fig.~\ref{fig:proj_cost}, we empirically show that a transformer model minimizes the projection loss $J_{\text{proj}}$ during training. Here, we train a vision transformer \cite{dosovitskiy2021an}, ViT-tiny model in particular, on the ImageNet-1K classification task and compute the average of $J_{\text{proj}}$ across attention heads and layers. This result suggests that during training, the transformer learns to perform PCA at each self-attention layer by implicitly minimizing the projection loss $J_{\text{proj}}$. Thus, the value vector $\vv_j$, $j=1,\dots,N$, in self-attention layers converge to the values specified in Theorem~\ref{theorem:attenion-pca}. We provide more details on the computation of $J_{\text{proj}}$ in Appendix~\ref{appx:J_proj}. 
\subsubsection{Learning Eigenvectors of \texorpdfstring{$\widetilde{\mK}_{\vvarphi}$}\  in Eqn.~(\ref{eqn:eig_K})}\label{subsubsec:learning_eg}
In this section, we study empirical results that confirm Eqn.~(\ref{eqn:eig_K}). In particular, we aim to verify that after training, the value matrix $\mV:=[\vv_1,\cdots,\vv_N]^\top$ captures the eigenvectors $\va_d$, $d=1,\dots,D_v$, of the Gram matrix $\widetilde{\mK}_{\vvarphi}$. 

\begin{figure*}[t!]
  \centering  \includegraphics[width=1.0\linewidth]{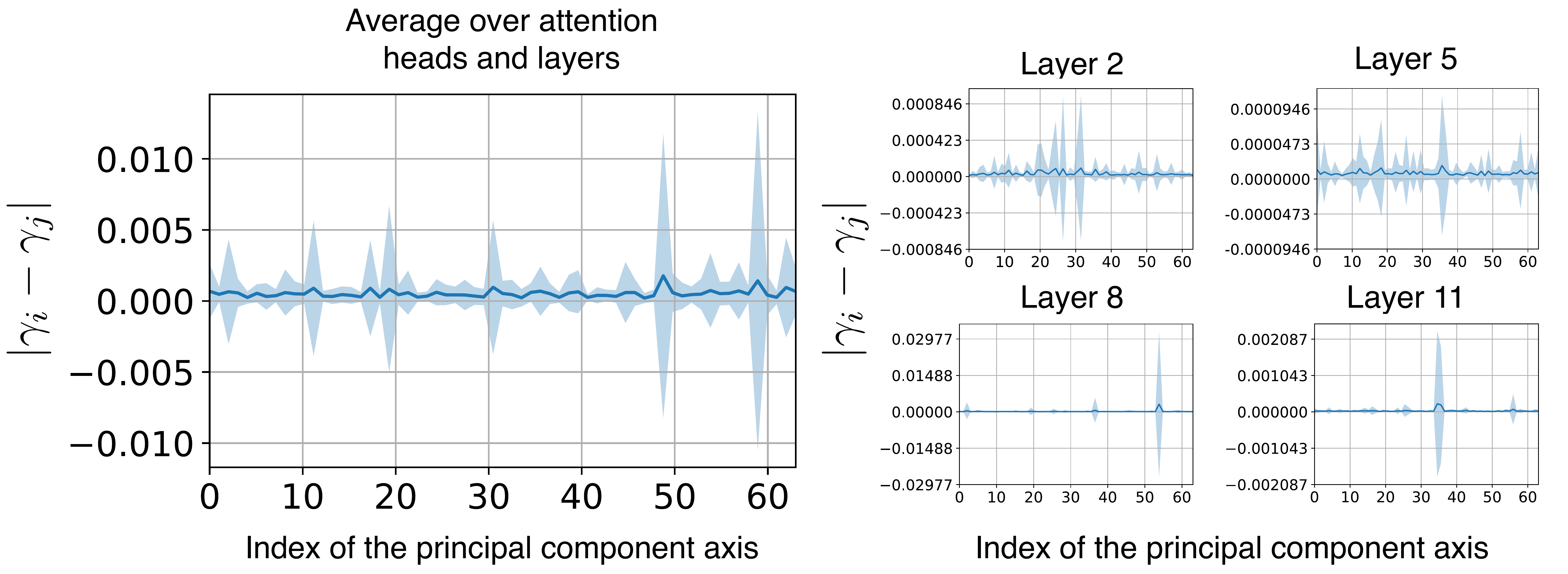}
  \caption{\small Mean and standard deviation of the absolute differences of elements in the constant vector $\mathbf{1}\lambda_d$, $d=1,\dots,D_v$. The mean should be $0$ with small standard deviations when $v_{dj}$ are close to the values predicted in Theorem~\ref{theorem:attenion-pca}. For comparison, we observe  that the max, min, mean, and median of the absolute values of all the eigenvalues, averaged over all attention heads and layers, are 648.46, 4.65, 40.07, and 17.73, respectively, which are much greater than the values of $|\gamma_i - \gamma_j|$.}
  \label{fig:eigenval_diff}
\end{figure*}

We first compute $\va_d$ in terms of $\mV$. Recall from Eqn.~(\ref{eqn:eig_K}) that $\va_d = [a_{d1},a_{d2},\cdots,a_{dN}]^\top$. We denote the diagonal matrix $\mG:=\text{diag}(1/g(\vk_1),\dots,1/g(\vk_N))$, the matrix $\mA := [\va_{1}, \dots, \va_{D_v}]$, and rewrite the value vectors $\vv_j$, $j=1,\dots,N$, as follows:
{\small
\begin{align}
\vv_j &= \Big[\frac{\va_{1}[j]}{g(\vk_j)} - \frac{1}{N}\sum_{j'=1}^N\frac{\va_{1}[j']}{g(\vk_{j})},\dots, \frac{\va_{D_{v}}[j]}{g(\vk_j)} - \frac{1}{N}\sum_{j'=1}^N\frac{\va_{D_{v}}[j']}{g(\vk_{j})}\Big]. \nonumber 
\end{align} }The value matrix $\mV$ in self-attention is then given by
\begin{align}
\mV &= \mG \mA - \mG\textbf{1}_{N}\mA \Leftrightarrow \mA = (\mI - \textbf{1}_{N})^{-1}\mG^{-1} \mV \nonumber 
\end{align}
Thus, given the value matrix $\mV=\mX\mW_V^\top$ that the self-attention learns after training, the estimation $\hat{\va}_d$ of $\va_d$ can be computed as
\begin{align}
\label{eqn:compute_ad}
\hat{\va}_d &= (\mI - \textbf{1}_{N})^{-1}\mG^{-1}\mV[:,d]. \nonumber
\end{align}
We empirically verify that $\frac{\widetilde{\mK}_{\vvarphi}\hat{\va}_d}{N\hat{\va}_d} = \vgamma = [\gamma_1,\dots,\gamma_N]$ where $\gamma_1 = \dots = \gamma_N = \text{const}$, which confirms that $\hat{\va}_d$ is an eigenvector of $\widetilde{\mK}_{\vvarphi}$. In particular, in Fig.~\ref{fig:eigenval_diff}(Left), we plot the average pairwise absolute differences of $\gamma_i$ and $\gamma_j$, $i \ne j$, $i,j = 1,\dots,N$, for each principal component axis of $\widetilde{\mK}_{\vvarphi}$. The results are averaged over all attention heads and all layers in the 12-layer ViT-tiny model trained on ImageNet-1K. As can be seen in our figure, the absolute difference between any pair of $\gamma_i$ and $\gamma_j$ is almost 0 with a very small standard deviation. Similar results are observed at each layer when averaging over all attention heads in that layer. In Fig.~\ref{fig:eigenval_diff}(Right), we show the results for Layers 2, 5, 8, and 11 in the model. For comparison, we observe  that the max, min, mean, and median of the absolute values of these $D_v$ eigenvalues, averaged over all attention heads and layers, are 648.46, 4.65, 40.07, and 17.73, respectively, which are much greater than the values of $|\gamma_i - \gamma_j|$. These results empirically justify that $\frac{\widetilde{\mK}_{\vvarphi}\hat{\va}_d}{N\hat{\va}_d} = \text{const}$ and the value matrix $\mV$ captures the eigenvectors of $\widetilde{\mK}_{\vvarphi}$ after the transformer model is trained, as suggested in Eqn.~(\ref{eqn:eig_K}).

In order to prove that after training a transformer with the value vectors $\vv_j$ parameterized as $\mW_V\vx_j$, $j=1,\dots,N$, using stochastic gradient descent, $v_{dj}$ converges to $\frac{a_{dj}}{g(\vk_j)} - \frac{1}{N}\sum_{j'=1}^N\frac{a_{dj'}}{g(\vk_{j})}$ as stated in Theorem~\ref{theorem:attenion-pca}, it is sufficient to prove that after the training, the outputs $\vh_{i}$ of self-attention become projections of the query vectors $\vq_{i}$, $i=1,\dots,N$, onto $D_v$ principal component axes in the feature space $\vvarphi$, i.e., the eigenvectors $\vu_1,\dots,\vu_{D_v}$ of the covariance matrix $\mC$. To theoretically prove this result for a multi-layer multi-head softmax transformer trained to explicitly minimize a particular loss, e.g., cross-entropy or L$2$ loss, using stochastic gradient descent is indeed challenging due to the highly nonconvex nature of the optimization problem and the nonlinear structure of the model. Our experimental results in Section~\ref{subsubsec:J_proj} and~\ref{subsubsec:learning_eg} empirically justify this result and serve as guidance for the theoretical proof, which we leave for future work.

\section{Robust Softmax Attention} 
In this section, we employ our kernel PCA framework in Section~\ref{sec:kpca-attn} to derive a new class of robust attention, namely, \emph{Attention with Robust Principal Components} (RPC-Attention). It is a well-known problem that both PCA and kernel PCA are sensitive to grossly corrupted data routinely encountered in modern applications~\cite{candes2009robust,nguyen2008robust,hubert2005robpca,de2001robust}. Since self-attention performs kernel PCA by projecting the query vectors $\vq_{i}$, $i=1,\dots,N$, onto principal components in a feature space as derived in Section~\ref{sec:deriving-attn}, it is also vulnerable to data corruption and perturbation. Our RPC-Attention robustifies self-attention by solving a convex program known as Principal Component Pursuit (PCP)~\cite{candes2009robust}.
\subsection{Principal Component Pursuit} 
Given corrupted measurement matrix $\mM \in \RR^{N\times D}$, both PCA and PCP aim to recover a low-rank matrix $\mL \in \RR^{N\times D}$ from $\mM$. However, while PCA models the corruption by a small Gaussian noise term, PCP models the corruption by a matrix $\mS \in \RR^{N\times D}$ that can have arbitrarily large magnitude with sparse supports. In particular, PCP solves the following convex optimization problem:
\begin{equation*}
\label{eq:sdp}
  \begin{array}{l}
    \text{minimize}_{\mL, \mS} \quad \|\mL\|_* + \lambda \|\mS\|_1 \quad
    \text{subject to} \quad \mL + \mS = \mM, 
  \end{array}
\end{equation*}
where $\|\mL\|_*$ is the nuclear norm of $\mL$, i.e., the sum of the singular values of $\mL$, and $\|\mS\|_1=\sum_{id}|S_{id}|$ is the $\ell_1$-norm of $\mS$. From~\cite{candes2009robust}, under minimal assumptions on the rank and sparsity of $\mL$ and $\mS$, the PCP solution exactly recovers the low-rank component $\mL$ and the sparse component $\mS$. \emph{Since PCP can recover the principal components of a data matrix even when a positive fraction of the measurements are arbitrarily corrupted, it is more robust than PCA}.
\subsection{Attention with Robust Principal Components} 
In self-attention, following our kernel PCA framework in Section~\ref{sec:kpca-attn}, the dataset $M$ is given as $M = \{\vk_1,\dots,\vk_N\}\subset \RR^{D}$ and $\vk_1,\dots,\vk_N$ are key vectors. Thus, the key matrix $\mK:=[\vk_1,\cdots,\vk_N]^\top \in \RR^{N\times D}$ in self-attention can be set as the measurement matrix $\mM$ in PCP. Then, the PCP for self-attention can be formulated as
\begin{equation}
\label{eq:sdp-attn}
  \begin{array}{l}
    \text{minimize}_{\mL, \mS} \quad \|\mL\|_* + \lambda \|\mS\|_1 \quad
    \text{subject to} \quad \mL + \mS = \mK. 
  \end{array}
\end{equation}
Following~\cite{candes2009robust}, we utilize the Alternating Direction Method of Multipliers (ADMM) algorithm introduced in~\cite{lin2010augmented,yuan2009sparse} to solve the convex program~(\ref{eq:sdp-attn}). The augmented Lagrangian of~(\ref{eq:sdp-attn}) is 
\begin{align*}
    \mathcal{L}(\mL, \mS, \mY)=\|\mL\|_*+\lambda \|\mS\|_1+
    \langle \mY, \mK - \mL - \mS \rangle + \mu/2 \|\mK - \mL - \mS\|_F^2.
\end{align*}
An ADMM solves the convex program~(\ref{eq:sdp-attn}) by iterating the following steps until convergence: [1] setting $\mS_{k+1}=\arg\min_{\mS} \mathcal{L}(\mL_k, \mS, \mY_k)$, [2] setting $\mL_{k+1}=\arg\min_{\mL} \mathcal{L}(\mL, \mS_{k+1}, \mY_k)$, and [3] updating the Lagrange multiplier matrix $\mY_{k+1} = \mY_k+\mu (\mK - \mL_{k+1} - \mS_{k+1})$. We define $\mathcal{S_\tau}(x):=\text{sgn}(x)\max(|x|-\tau,0)$ as an element-wise shrinkage operator and $\mathcal{D_\tau}(\mX)=\mU\mathcal{S_\tau}(\Sigma)\mV^*$ as a singular value thresholding operator with the singular value decomposition of $\mX=\mU \Sigma \mV^*$. As proven in~\cite{candes2009robust}, we can rewrite steps [1] and [2] as
\begin{align*}
    \arg\min_{\mS} \mathcal{L}(\mL, \mS, \mY) = \mathcal{S_{\lambda / \mu}}(\mK - \mL + \mu^{-1}\mY);\, \arg\min_{\mL} \mathcal{L}(\mL, \mS, \mY) = \mathcal{D}_{\mu}(\mK - \mS - \mu^{-1}\mY).
\end{align*}
$\mathcal{D_\mu}$ finds a low-rank approximation of $\mK - \mS - \mu^{-1}\mY$. Thus, we obtain an approximation of the above equation by replacing $\mathcal{D_\mu}$ by a low-rank approximation operator. Such 
an approximation takes a
\begin{algorithm}[t!]
\begin{algorithmic}
{\small
    \STATE \textbf{initialize:} $\mS_0 = \mY_0 = \textbf{0}$; $\mu, \lambda > 0$.
    \WHILE{not converged}
      \STATE compute $\mS_{k+1} = \mathcal{\mS_{\lambda/\mu}}(\mK - \mL_k + \mu^{-1}\mY_k)$;
      \STATE compute $\mL_{k+1} = \text{Softmax}(\mK - \mS_{k+1} - \mu^{-1}\mY_k, \mK - \mS_{k+1} - \mu^{-1}\mY_k)$;
      \STATE compute $\mY_{k+1} = \mY_k + \mu(\mK - \mL_{k+1} - \mS_{k+1})$;
    \ENDWHILE
    \STATE \textbf{output:} $\mL$.}
  \end{algorithmic}
 \caption{\small Principal Attention Pursuit (PAP)}
\label{algo:pap}
\end{algorithm}
step towards the minimum of $\mathcal{L}(\mL, \mS, \mY)$ when fixing $\mS$ and $\mY$. 
It has been empirically observed and theoretically proven that the output matrix $\mH$ of self-attention is low-rank, a phenomenon known as over-smoothing or rank collapse~\cite{shi2022revisiting,wang2022antioversmooth,dong2021attention}. Therefore, we can replace $\mathcal{D_\mu}$ by a self-attention operator. The ADMM method applied to self-attention, which we name \emph{Principal Attention Pursuit} (PAP), is given by Algorithm~\ref{algo:pap}. We define our RPC-Attention as 
\begin{definition}[Attention with Robust Principal Components]
\label{def:rpc-attn}
\vspace{-0.05in}
An RPC-Attention performs the PAP in Algorithm~\ref{algo:pap} for $n$ iterations with $\lambda$ as a hyperparameter. For the key matrix $\mK\in \RR^{N \times D}$, RPC-Attention sets $\mu = ND /4\|\mK\|_1$ as suggested in \cite{candes2009robust}, where $\|\mK\|_1=\sum_{id}|K_{id}|$. The output matrix $\mH$ of RPC-Attention is set to be the low-rank output matrix $\mL$ from PAP.
\end{definition}
\section{Experimental Results}
We aim to numerically show that: (i) RPC-Attention achieves competitive or even better accuracy than the baseline softmax attention on clean data, and (ii) the advantages of RPC-Attention are more prominent when there is a contamination of samples across different types of data and a variety of tasks. We also validate the performance of the Scaled Attention proposed in Remark~\ref{rm:parameterization}. 

Throughout our experiments, we compare the
performance of our proposed models with the baseline softmax attention of the same configuration. All of our results are averaged over 5 runs with different seeds and run on 4 A100 GPU. Details on the models and training settings are provided in
Appendix~\ref{appx:exp_details} and additional experimental results are provided in Appendix~\ref{appx:extra_experiments}. Primarily, we focus on a ViT-tiny model backbone~\cite{dosovitskiy2021an}, but included in the appendix are experiments on a larger model backbone, ViT-\rach{base}, and a state of the art (SOTA) robust model, Fully Attentional Networks (FAN) ~\cite{zhou2022understanding}.
\subsection{Vision Tasks: ImageNet-1K Object Classification}

We implement PAP in Algorithm~\ref{algo:pap} in the symmetric softmax attention layers of a ViT-tiny model and compare it to the standard symmetric model as our baseline. We refer to our model as RPC-SymViT and the baseline model as SymViT. For RPC-SymViT, we study two settings. In the first setting, which we denote by RPC-SymViT ($n$iter/layer1), $n=4,5,6$, to maintain the computational efficiency of the model, we apply $n$ PAP iterations only at the first layer to recover a cleaner data matrix that is then sent to the subsequent layers in the model. In the second setting, which we denote by RPC-SymViT ($n$iter/all-layer), $n=2$, we apply $n$ PAP iterations at all layers. 
We note that an iterative scheme has the potential to have an increased computational load, hence, we provide a comparison on the number of flops per sample, run time per sample, memory and number of parameters between RPC-SymViT and the baseline in Appendix \ref{appx:eff}, showing a comparable efficiency. 

{\bf Robustness against Data Corruptions.}
\begin{table*}[!t]
{\footnotesize	
  \begin{center}
    \caption{\small Top-1, Top-5 accuracy (\%) , mean corruption error (mCE), and area under the precision-recall curve (AUPR) of RPC-SymViT and SymViT on clean ImageNet-1K data and popular standard robustness benchmarks for image classification. RPC-SymViT ($n$iter/layer1) applies $n$ PAP iterations only at the first layer. RPC-SymViT ($n$iter/all-layer) applies $n$ PAP iterations at all layers.} 
    \label{tab:table1}
    \begin{tabular}{lcccccccc} 
    \toprule
      \multirow{2}{*}{Model} & \multicolumn{2}{c}{IN-1K} & IN-R & IN-A & \multicolumn{2}{c}{IN-C} & IN-O \\
      & Top-1 $\uparrow$ & Top-5 $\uparrow$ & Top-1 $\uparrow$ & Top-1 $\uparrow$ & Top-1 $\uparrow$ & mCE $\downarrow$ & AUPR $\uparrow$\\
        \midrule
      SymViT (baseline) & 70.44 & 90.17 & 28.98 & 6.51 & 41.45 & 74.75 & 17.43 \\
      \midrule
      RPC-SymViT (4iter/layer1) & 70.94 & 90.47 & 29.99 & 6.96 & 42.35 & 73.58 & 19.32 \\
      RPC-SymViT (5iter/layer1) & 71.31 & 90.59 & \textbf{30.28} & 7.27 & 42.43 & 73.43 & \textbf{20.35} \\
      RPC-SymViT (6iter/layer1) & \textbf{71.49} & \textbf{90.68} & 30.03 & 7.33 & \textbf{42.76} & \textbf{73.03} & 20.29\\
      \midrule
      RPC-SymViT (2iter/all-layer) & 70.59 & 90.15 & 29.23 & \textbf{7.55} & 41.64 & 74.52 & 19.18 \\
    \bottomrule
    \end{tabular}
  \end{center}
}
\end{table*}
\begin{table*}[!t]
{\footnotesize	
\setlength{\tabcolsep}{4.7 pt} 
  \begin{center}
    \caption{\small Top-1/5 accuracy (\%) on attacked ImageNet-1K data. RPC-SymViT ($n$iter/layer1) applies $n$ PAP iterations at the first layer. RPC-SymViT ($n$iter/all-layer) applies $n$ PAP iterations at all layers.}
    \label{tab:table3}
   \begin{tabular}{ccccccc} 
    \toprule
      \multirow{2}{*}{Attack} & \multirow{2}{*}{Metric/Model}&  
      SymViT & RPC-SymViT & RPC-SymViT & RPC-SymViT & RPC-SymViT \\
      & & (baseline) & (4iter/layer1) & (5iter/layer1) & (6iter/layer1) & (2iter/all-layer) \\
      \midrule
      \multirow{2}{*}{PGD} & Top-1 $\uparrow$ & 4.98 & 5.15&5.11& 5.20& \textbf{6.12}\\
      & Top-5 $\uparrow$ 
      & 10.41  & 11.20  & 11.13  & 11.34  & \textbf{13.24} \\
      \midrule
      \multirow{2}{*}{FGSM} & Top-1 $\uparrow$ & 23.38  & 26.62  & 26.75 & 27.22 & \textbf{29.20} 
      \\
      & Top-5 $\uparrow$ & 53.82 & 56.87 & 57.19 & 57.55 & \textbf{59.63} 
      \\
      \midrule
      \multirow{2}{*}{SPSA} & Top-1 $\uparrow$ & 47.94 & 48.13 & 49.29 & 48.75 & \textbf{51.01}
      \\
      & Top-5 $\uparrow$ & 82.63 & 82.87 & 83.52 & 83.27 & \textbf{83.66}
      \\
      \midrule
      \multirow{2}{*}{SLD} & Top-1 $\uparrow$ & 67.91 & 68.48 & 68.60 & \textbf{68.99} & 67.60 
      \\
      & Top-5 $\uparrow$ & 89.68 & 90.16 & 90.18 & \textbf{90.38} & 89.79
      \\
      \midrule
      \multirow{2}{*}{CW} & Top-1 $\uparrow$ & 48.44 & 50.00 & 50.08 & \textbf{50.36} & 48.77 
      \\
      & Top-5 $\uparrow$ & 72.68 & 74.24 & 74.14 & \textbf{74.46} & 72.91
      \\
      \midrule
      \multirow{2}{*}{Noise} & Top-1 $\uparrow$ & 67.81 & 68.37 & 68.72 & \textbf{68.81} & 68.09 
      \\
      & Top-5 $\uparrow$ & 88.51 & 89.00 & 89.13 & \textbf{89.27} & 88.64
      \\
      \midrule
      \multirow{2}{*}{AutoAttack} & Top-1 $\uparrow$ &  23.09  & 24.56 & 24.68 & \textbf{24.74} & 23.51
      \\
      & Top-5 $\uparrow$ & 63.48 & 65.11 & 65.09 & \textbf{65.13} & 64.06
      \\
      \bottomrule
    \end{tabular}
  \end{center}
}
\end{table*}
\begin{figure}[!t]
  \centering  \includegraphics[width=1.0\linewidth]{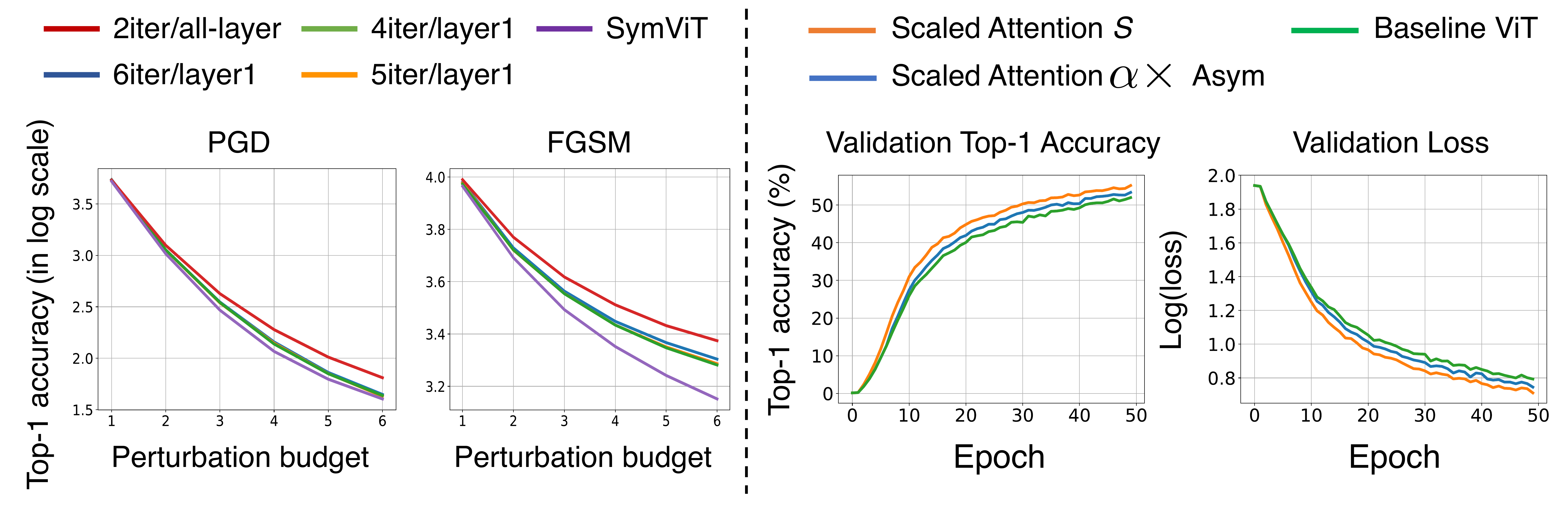}
  \caption{\small \textbf{Left:} Top-1 accuracy of RPC-SymViT vs.\ baseline SymViT evaluated on PGD/FGSM attacked  ImageNet-1K validation set across increasing perturbation budgets. \textbf{Right:} Validation top-1 accuracy (\%) and loss of Scaled Attention vs.\ the baseline asymmetric softmax attention in ViT for the first 50 training epochs. } 
  \label{fig:attack}
\end{figure}
To benchmark robustness to data corruptions, we use the standard datasets, ImageNet-C (IN-C)~\cite{DBLP:journals/corr/abs-1903-12261}, ImageNet-A (IN-A), ImageNet-O (IN-O)~\cite{hendrycks2021natural}, and ImageNet-R (IN-R)~\cite{DBLP:journals/corr/abs-2006-16241}. We provide details on each dataset and the metrics for evaluation in Appendix~\ref{appx:robust_corrupt_details}. The direction of increasing or decreasing values of these metrics signifying greater robustness are indicated in the Table~\ref{tab:table1} with an arrow, along with the results on each dataset. 

Across all evaluations, RPC-SymViT outperforms the SymViT baseline, thereby justifying the advantages of our RPC-Attention. Particularly, RPC-SymViT with 6 iterations in the 1st layer achieves an improvement of over 1\% in terms of accuracy on the clean ImageNet-1K validation set and almost 3 AUPR on ImageNet-O compared to the SymViT. This result is consistent with the intuition that a higher number of iterations executed on a consistent data matrix leads to cleaner data. The full details of all corruption types are presented in Appendix~\ref{appx:allimC}.   

{\bf Robustness under Adversarial Attacks.} We report the top-1 and top-5 accuracy of the models on ImageNet-1K distorted by white box attacks, including PGD~\cite{madry2018towards}, FGSM~\cite{43405}, SLD~\cite{DBLP:journals/corr/abs-1904-13000}, and CW~\cite{DBLP:journals/corr/CarliniW16a} attacks. We further examine the models on the black-box attack SPSA~\cite{uesato2018adversarial}, AutoAttack~\cite{croce2020reliable}, and a noise-adding attack. 
In Table~\ref{tab:table3}, we present the results of all attacks, where PGD and FGSM are reported for the maximum perturbation budget. 

On all attacks, RPC-SymViT outperforms SymViT by a substantial margin, demonstrating the effectiveness of RPC-Attention. 
Notably, as the square attack in AutoAttack is also a score-based black box attack, in order to further verify our method, we include the results of RPC-SymViT (6iter/layer1) and the baseline SymViT on this attack in the Appendix~\ref{appx:square} in which our model also performs better. This result together with our model's considerable improvement over the baseline on the black-box SPSA attack justifies that RPC-Attention's robustness against adversarial attacks is not due to any form of gradient masking. In addition, we illustrate RPC-SymViT's robustness across increasing perturbations for PGD and FGSM in Fig.~\ref{fig:attack}(left). Details on the evaluation of the models under all attacks are provided in Appendix \ref{appx:attack}.

{\bf ADE20K Image Segmentation.} We continue evaluating the benefits of our method by implementing RPC-Attention in a Segmenter model~\cite{strudel2021segmenter} and providing results on the ADE20K image segmentation task \cite{8100027}. Table \ref{tab:table4} in Appendix~\ref{app:ade20k} shows that Segmenter with the RPC-SymViT backbone outperforms the Segmenter with the baseline SymViT backbone on both clean and corrupted data.



\subsection{Language Tasks: WikiText-103 Language Modeling}
We assess our model on the large-scale WikiText-103 language modeling task~\cite{DBLP:journals/corr/MerityXBS16}. 
Using a standard transformer language model \cite{vaswani2017attention}, with a symmetric attention layer (Symlm), we replace it with an RPC-Attention layer (RPC-Symlm). As with RPC-SymViT, we only implement RPC-Attention in the first four layers of Symlm and run PAP for 4 iterations to save on computational overhead. 
 The results of the validation and test perplexity (PPL) summarized in Table \ref{tab:table5} validate the advantage of our method, surpassing the baseline by at least 1 PPL for all datasets.

\subsection{Validating the Benefits of Scaled Attention}

\begin{table}[t!]
{\footnotesize	
  \begin{center}
\caption{\small Validation/test perplexity (PPL) on clean WikiText-103 and word swap attacked dataset. RPC-Symlm ($n$iter/layer$n_1$-$n_2$) applies $n$ iterations of PAP only in layers $n_1$ to $n_2$ of the model. }
    \label{tab:table5}
\vspace{0.4em}
    \begingroup

    \renewcommand{\arraystretch}{1} 
    \begin{tabular}{lcccc} 
    \toprule
      \multirow{2}{*}{Model} & \multicolumn{2}{c}{Wikitext-103} &  \multicolumn{2}{c}{Attacked Wikitext-103} \\
      & Valid-ppl $\downarrow$ & Test-ppl $\downarrow$ & Valid-ppl $\downarrow$ & Test-ppl $\downarrow$  \\
        \hline
      Symlm & 38.37 & 36.36 & 42.90 & 44.32 \\
      \midrule
      RPC-Symlm &\multirow{2}{*}{\textbf{37.38}} & \multirow{2}{*}{\textbf{35.26}} & \multirow{2}{*}{\textbf{41.83}} & \multirow{2}{*}{\textbf{43.16}}\\
      (4iter/layer1-4) & \\
    \bottomrule
    \end{tabular}
    \endgroup
  \end{center}
}
\end{table}

In this section, we provide an empirical comparison between the Scaled Attention in Remark~\ref{rm:parameterization} and the softmax attention. We train an asymmetric ViT-tiny model with two different versions of Scaled Attention. While the exact value of $\mS$ can be mathematically formulated as in Remark~\ref{rm:parameterization}, it might lead to numerical errors that are difficult to handle. Therefore, in the first version of Scaled Attention, we let $\mS$ be a learned parameter matrix in each layer (Scaled Attention $S$ in Fig.~\ref{fig:attack}(right)). In the second version of Scaled Attention (Scaled Attention $\alpha \times \text{Asym}$ in Fig.~\ref{fig:attack}(right)), given that $s_{jj'}=g(k_{j'})/g(k_{j})$, we rewrite $\mS$ as the product of a symmetric softmax attention matrix and the reciprocal of its transpose. The model learns this reciprocal by a learnable scalar $\alpha$ and we let $\mS=\alpha \mA_{Sym}$, where $\mA_{Sym}$ is the symmetric softmax attention. More details are in Appendix~\ref{appx:reparam}.

Fig.~\ref{fig:attack}(right) shows the top-1 validation accuracy and loss over 50 epochs when training ViT models equipped with Scaled Attention and softmax attention on the ImageNet-1K object classification task. The full training curve can also be found in Appendix~\ref{appx:reparam}. The results suggest that both versions of Scaled Attention outperform the softmax attention. This provides further evidence that self-attention learns to approximate a kernel PCA since the Scaled Attention with a more explicit structure of the value matrix $\mV$ suggested in Theorem~\ref{theorem:attenion-pca} obtains better performance than softmax attention.
\section{Related Works}
{\bf Theoretical Perspectives of Attention Mechanisms.} The study of the attention mechanism in transformers through different theoretical frameworks has been expanding. \cite{tsai-etal-2019-transformer} shows that attention can be analyzed as a kernel smoother over the inputs using an appropriate kernel score that is the similarities between them. \cite{choromanski2021rethinking,DBLP:journals/corr/abs-2006-04768,DBLP:journals/corr/abs-2006-16236,nguyen2021fmmformer}
reduce the quadratic complexity of transformers by linearizing the softmax kernel to improve the computational and memory efficiency. 
In addition, there are works interpreting transformers using the frameworks of ordinary/partial differential equations \cite{nguyen2023mitigating,nguyen2024pidformer,lu2019understanding,DBLP:journals/corr/abs-2110-11773,geshkovski2024emergence,geshkovski2023mathematical} and from probabilistic viewpoints with Gaussian Mixture Models \cite{nguyen2022improving,NEURIPS2021_23937b42,tang2021probabilistic,nguyen2022fourierformer,zhang-feng-2021-modeling-concentrated}. \cite{chen2023primalattention} provides a new perspective by emphasizing the asymmetry of the softmax kernel and recovers the self-attention mechanism from an asymmetric Kernel Singular Value Decomposition (KSVD) using the duality of the optimization problem. Another related work views transformers from the perspective of Support Vector Machines \cite{nguyen2023a,tarzanagh2023transformers}. We discuss~\cite{chen2023primalattention} and~\cite{tarzanagh2023transformers}, as well as the approaches that use matrix decomposition and iterative algorithms in deep models, in Appendix~\ref{appx:related-works}. Separate from these
works, our kernel PCA perspective derives softmax attention as a projection of the query vectors in a feature space. Using our framework, we are able to predict the exact explicit form of the value matrix in self-attention, demonstrating that this matrix captures the eigenvectors of the Gram matrix of the
key vectors in a feature space. Our work is the first to show this insight.

{\bf Robustness of Transformers.} There have been many works studying the robustness of Vision Transformers (ViT)~\cite{dosovitskiy2021an} against different types of attacks \cite{DBLP:journals/corr/abs-2103-14586,DBLP:journals/corr/abs-2105-07581,subramanya2022backdoor,zhou2022understanding}. Recent work that serves to address this include \cite{mao2022robust}, whereby new training strategies and architectural adjustments are proposed to improve the robustness of ViT.  In addition, \cite{nielsen2024elliptical} suggests employing a Mahalanobis distance metric to calculate attention weights, expanding the feature space along directions of high contextual relevance, thereby enhancing the model's robustness. Also, \cite{han2024designing} adapts traditional robust kernel density estimation techniques to create new classes of transformers that are resilient to adversarial attacks and data contamination. \cite{chen2023calibrating,bui2024revisiting} integrate a Gaussian process into attention for out-of-distribution detection, and \cite{tran2024equivariant} develops equivariant neural functional networks for transformers.  Our RPC-Attention is orthogonal to these methods.

\section{Concluding Remarks}
In this paper, we derive self-attention from kernel principal component analysis (kernel PCA) as a projection of the query vectors onto the principal component axes of the key matrix in a feature space. 
Using our kernel PCA framework, we derive a new class of robust attention, namely the Attention with Robust Principal Components (RPC-Attention), that is resilient to data contamination. A limitation of RPC-Attention is its derivation from an iterative algorithm that leads to its unrolling architecture, increasing the computational cost of the model. In our paper, we mitigate this by only replacing softmax attention with RPC-Attention in the first layer of the model and demonstrate that doing so is sufficiently effective for robustness. In addition, we provide a comparison of the efficiency of RPC-Attention with softmax attention in Appendix \ref{appx:eff} and find that we are comparable across all metrics at test time while only slightly less efficient during training. 
It is also interesting to extend our kernel PCA framework to explain multi-layer transformers. We leave these exciting directions as future work. 

\newpage

\begin{ack}
This research / project is supported by the National Research Foundation Singapore under the AI Singapore Programme (AISG Award No: AISG2-TC-2023-012-SGIL). This research / project is supported by the Ministry of Education, Singapore, under the Academic Research Fund Tier 1 (FY2023) (A-8002040-00-00, A-8002039-00-00). This research / project is also supported by the NUS Presidential Young Professorship Award (A-0009807-01-00).
\end{ack}

\bibliography{main}

\begin{thebibliography}{10}

\bibitem{al2019character}
Rami Al-Rfou, Dokook Choe, Noah Constant, Mandy Guo, and Llion Jones.
\newblock Character-level language modeling with deeper self-attention.
\newblock In {\em Proceedings of the AAAI Conference on Artificial
  Intelligence}, volume~33, pages 3159--3166, 2019.

\bibitem{amos2017optnet}
Brandon Amos and J~Zico Kolter.
\newblock Optnet: Differentiable optimization as a layer in neural networks.
\newblock In {\em International conference on machine learning}, pages
  136--145. PMLR, 2017.

\bibitem{andriushchenko2020square}
Maksym Andriushchenko, Francesco Croce, Nicolas Flammarion, and Matthias Hein.
\newblock Square attack: a query-efficient black-box adversarial attack via
  random search.
\newblock In {\em European conference on computer vision}, pages 484--501.
  Springer, 2020.

\bibitem{baevski2018adaptive}
Alexei Baevski and Michael Auli.
\newblock Adaptive input representations for neural language modeling.
\newblock In {\em International Conference on Learning Representations}, 2019.

\bibitem{bahdanau2014neural}
Dzmitry Bahdanau, Kyunghyun Cho, and Yoshua Bengio.
\newblock Neural machine translation by jointly learning to align and
  translate.
\newblock In {\em International Conference on Learning Representations}, 2015.

\bibitem{bai2019deep}
Shaojie Bai, J~Zico Kolter, and Vladlen Koltun.
\newblock Deep equilibrium models.
\newblock {\em Advances in neural information processing systems}, 32, 2019.

\bibitem{DBLP:journals/corr/abs-2103-14586}
Srinadh Bhojanapalli, Ayan Chakrabarti, Daniel Glasner, Daliang Li, Thomas
  Unterthiner, and Andreas Veit.
\newblock Understanding robustness of transformers for image classification.
\newblock In {\em Proceedings of the IEEE/CVF international conference on
  computer vision}, pages 10231--10241, 2021.

\bibitem{bishop2006pattern}
Christopher~M Bishop and Nasser~M Nasrabadi.
\newblock {\em Pattern recognition and machine learning}, volume~4.
\newblock Springer, 2006.

\bibitem{NEURIPS2020_1457c0d6}
Tom Brown, Benjamin Mann, Nick Ryder, Melanie Subbiah, Jared~D Kaplan, Prafulla
  Dhariwal, Arvind Neelakantan, Pranav Shyam, Girish Sastry, Amanda Askell,
  Sandhini Agarwal, Ariel Herbert-Voss, Gretchen Krueger, Tom Henighan, Rewon
  Child, Aditya Ramesh, Daniel Ziegler, Jeffrey Wu, Clemens Winter, Chris
  Hesse, Mark Chen, Eric Sigler, Mateusz Litwin, Scott Gray, Benjamin Chess,
  Jack Clark, Christopher Berner, Sam McCandlish, Alec Radford, Ilya Sutskever,
  and Dario Amodei.
\newblock Language models are few-shot learners.
\newblock In {\em Advances in Neural Information Processing Systems},
  volume~33, pages 1877--1901. Curran Associates, Inc., 2020.

\bibitem{bui2024revisiting}
Long~Minh Bui, Tho~Tran Huu, Duy Dinh, Tan~Minh Nguyen, and Trong~Nghia Hoang.
\newblock Revisiting kernel attention with correlated gaussian process
  representation.
\newblock In {\em The 40th Conference on Uncertainty in Artificial
  Intelligence}, 2024.

\bibitem{candes2009robust}
Emmanuel~J Cand{\`e}s, Xiaodong Li, Yi~Ma, and John Wright.
\newblock Robust principal component analysis?
\newblock {\em Journal of the ACM (JACM)}, 58(3):1--37, 2011.

\bibitem{DBLP:journals/corr/CarliniW16a}
Nicholas Carlini and David Wagner.
\newblock Towards evaluating the robustness of neural networks.
\newblock In {\em 2017 ieee symposium on security and privacy (sp)}, pages
  39--57. Ieee, 2017.

\bibitem{chen2021decision}
Lili Chen, Kevin Lu, Aravind Rajeswaran, Kimin Lee, Aditya Grover, Misha
  Laskin, Pieter Abbeel, Aravind Srinivas, and Igor Mordatch.
\newblock Decision transformer: Reinforcement learning via sequence modeling.
\newblock {\em Advances in neural information processing systems},
  34:15084--15097, 2021.

\bibitem{chen2023calibrating}
Wenlong Chen and Yingzhen Li.
\newblock Calibrating transformers via sparse gaussian processes.
\newblock In {\em The Eleventh International Conference on Learning
  Representations}, 2023.

\bibitem{chen2023primalattention}
Yingyi Chen, Qinghua Tao, Francesco Tonin, and Johan Suykens.
\newblock Primal-attention: Self-attention through asymmetric kernel svd in
  primal representation.
\newblock {\em Advances in Neural Information Processing Systems}, 36, 2024.

\bibitem{choromanski2021rethinking}
Krzysztof~Marcin Choromanski, Valerii Likhosherstov, David Dohan, Xingyou Song,
  Andreea Gane, Tamas Sarlos, Peter Hawkins, Jared~Quincy Davis, Afroz
  Mohiuddin, Lukasz Kaiser, David~Benjamin Belanger, Lucy~J Colwell, and Adrian
  Weller.
\newblock Rethinking attention with performers.
\newblock In {\em International Conference on Learning Representations}, 2021.

\bibitem{JMLR:v24:22-1144}
Aakanksha Chowdhery, Sharan Narang, Jacob Devlin, Maarten Bosma, Gaurav Mishra,
  Adam Roberts, Paul Barham, Hyung~Won Chung, Charles Sutton, Sebastian
  Gehrmann, Parker Schuh, Kensen Shi, Sasha Tsvyashchenko, Joshua Maynez,
  Abhishek Rao, Parker Barnes, Yi~Tay, Noam Shazeer, Vinodkumar Prabhakaran,
  Emily Reif, Nan Du, Ben Hutchinson, Reiner Pope, James Bradbury, Jacob
  Austin, Michael Isard, Guy Gur-Ari, Pengcheng Yin, Toju Duke, Anselm
  Levskaya, Sanjay Ghemawat, Sunipa Dev, Henryk Michalewski, Xavier Garcia,
  Vedant Misra, Kevin Robinson, Liam Fedus, Denny Zhou, Daphne Ippolito, David
  Luan, Hyeontaek Lim, Barret Zoph, Alexander Spiridonov, Ryan Sepassi, David
  Dohan, Shivani Agrawal, Mark Omernick, Andrew~M. Dai,
  Thanumalayan~Sankaranarayana Pillai, Marie Pellat, Aitor Lewkowycz, Erica
  Moreira, Rewon Child, Oleksandr Polozov, Katherine Lee, Zongwei Zhou, Xuezhi
  Wang, Brennan Saeta, Mark Diaz, Orhan Firat, Michele Catasta, Jason Wei,
  Kathy Meier-Hellstern, Douglas Eck, Jeff Dean, Slav Petrov, and Noah Fiedel.
\newblock Palm: Scaling language modeling with pathways.
\newblock {\em Journal of Machine Learning Research}, 24(240):1--113, 2023.

\bibitem{croce2020reliable}
Francesco Croce and Matthias Hein.
\newblock Reliable evaluation of adversarial robustness with an ensemble of
  diverse parameter-free attacks.
\newblock In {\em International conference on machine learning}, pages
  2206--2216. PMLR, 2020.

\bibitem{de2001robust}
Fernando De~la Torre and Michael~J Black.
\newblock Robust principal component analysis for computer vision.
\newblock In {\em Proceedings Eighth IEEE International Conference on Computer
  Vision. ICCV 2001}, volume~1, pages 362--369. IEEE, 2001.

\bibitem{devlin-etal-2019-bert}
Jacob Devlin, Ming-Wei Chang, Kenton Lee, and Kristina Toutanova.
\newblock {BERT}: Pre-training of deep bidirectional transformers for language
  understanding.
\newblock In Jill Burstein, Christy Doran, and Thamar Solorio, editors, {\em
  Proceedings of the 2019 Conference of the North {A}merican Chapter of the
  Association for Computational Linguistics: Human Language Technologies,
  Volume 1 (Long and Short Papers)}, pages 4171--4186, Minneapolis, Minnesota,
  June 2019. Association for Computational Linguistics.

\bibitem{dong2021attention}
Yihe Dong, Jean-Baptiste Cordonnier, and Andreas Loukas.
\newblock Attention is not all you need: Pure attention loses rank doubly
  exponentially with depth.
\newblock In {\em International Conference on Machine Learning}, pages
  2793--2803. PMLR, 2021.

\bibitem{dosovitskiy2021an}
Alexey Dosovitskiy, Lucas Beyer, Alexander Kolesnikov, Dirk Weissenborn,
  Xiaohua Zhai, Thomas Unterthiner, Mostafa Dehghani, Matthias Minderer, Georg
  Heigold, Sylvain Gelly, Jakob Uszkoreit, and Neil Houlsby.
\newblock An image is worth 16x16 words: Transformers for image recognition at
  scale.
\newblock In {\em International Conference on Learning Representations}, 2021.

\bibitem{NEURIPS2021_23937b42}
Prasad Gabbur, Manjot Bilkhu, and Javier Movellan.
\newblock Probabilistic attention for interactive segmentation.
\newblock In M.~Ranzato, A.~Beygelzimer, Y.~Dauphin, P.S. Liang, and J.~Wortman
  Vaughan, editors, {\em Advances in Neural Information Processing Systems},
  volume~34, pages 4448--4460. Curran Associates, Inc., 2021.

\bibitem{gal2023an}
Rinon Gal, Yuval Alaluf, Yuval Atzmon, Or~Patashnik, Amit~Haim Bermano, Gal
  Chechik, and Daniel Cohen-or.
\newblock An image is worth one word: Personalizing text-to-image generation
  using textual inversion.
\newblock In {\em The Eleventh International Conference on Learning
  Representations}, 2023.

\bibitem{geng2021attention}
Zhengyang Geng, Meng-Hao Guo, Hongxu Chen, Xia Li, Ke~Wei, and Zhouchen Lin.
\newblock Is attention better than matrix decomposition?
\newblock In {\em International Conference on Learning Representations}, 2021.

\bibitem{geshkovski2023mathematical}
Borjan Geshkovski, Cyril Letrouit, Yury Polyanskiy, and Philippe Rigollet.
\newblock A mathematical perspective on transformers.
\newblock {\em arXiv preprint arXiv:2312.10794}, 2023.

\bibitem{geshkovski2024emergence}
Borjan Geshkovski, Cyril Letrouit, Yury Polyanskiy, and Philippe Rigollet.
\newblock The emergence of clusters in self-attention dynamics.
\newblock {\em Advances in Neural Information Processing Systems}, 36, 2024.

\bibitem{43405}
Ian Goodfellow, Jonathon Shlens, and Christian Szegedy.
\newblock Explaining and harnessing adversarial examples.
\newblock In {\em International Conference on Learning Representations}, 2015.

\bibitem{han2024designing}
Xing Han, Tongzheng Ren, Tan Nguyen, Khai Nguyen, Joydeep Ghosh, and Nhat Ho.
\newblock Designing robust transformers using robust kernel density estimation.
\newblock {\em Advances in Neural Information Processing Systems}, 36, 2024.

\bibitem{DBLP:journals/corr/abs-2006-16241}
Dan Hendrycks, Steven Basart, Norman Mu, Saurav Kadavath, Frank Wang, Evan
  Dorundo, Rahul Desai, Tyler Zhu, Samyak Parajuli, Mike Guo, et~al.
\newblock The many faces of robustness: A critical analysis of
  out-of-distribution generalization.
\newblock In {\em Proceedings of the IEEE/CVF international conference on
  computer vision}, pages 8340--8349, 2021.

\bibitem{DBLP:journals/corr/abs-1903-12261}
Dan Hendrycks and Thomas Dietterich.
\newblock Benchmarking neural network robustness to common corruptions and
  perturbations.
\newblock In {\em International Conference on Learning Representations}, 2019.

\bibitem{hendrycks2021natural}
Dan Hendrycks, Kevin Zhao, Steven Basart, Jacob Steinhardt, and Dawn Song.
\newblock Natural adversarial examples.
\newblock In {\em Proceedings of the IEEE/CVF conference on computer vision and
  pattern recognition}, pages 15262--15271, 2021.

\bibitem{hewitt-liang-2019-designing}
John Hewitt and Percy Liang.
\newblock Designing and interpreting probes with control tasks.
\newblock In {\em Proceedings of the 2019 Conference on Empirical Methods in
  Natural Language Processing and the 9th International Joint Conference on
  Natural Language Processing (EMNLP-IJCNLP)}, pages 2733--2743, Hong Kong,
  China, November 2019. Association for Computational Linguistics.

\bibitem{hubert2005robpca}
Mia Hubert, Peter~J Rousseeuw, and Karlien Vanden~Branden.
\newblock Robpca: a new approach to robust principal component analysis.
\newblock {\em Technometrics}, 47(1):64--79, 2005.

\bibitem{janner2021offline}
Michael Janner, Qiyang Li, and Sergey Levine.
\newblock Offline reinforcement learning as one big sequence modeling problem.
\newblock {\em Advances in neural information processing systems},
  34:1273--1286, 2021.

\bibitem{jumper2021highly}
John Jumper, Richard Evans, Alexander Pritzel, Tim Green, Michael Figurnov,
  Olaf Ronneberger, Kathryn Tunyasuvunakool, Russ Bates, Augustin
  {\v{Z}}{\'\i}dek, Anna Potapenko, et~al.
\newblock Highly accurate protein structure prediction with alphafold.
\newblock {\em Nature}, 596(7873):583--589, 2021.

\bibitem{DBLP:journals/corr/abs-2006-16236}
Angelos Katharopoulos, Apoorv Vyas, Nikolaos Pappas, and Fran{\c{c}}ois
  Fleuret.
\newblock Transformers are rnns: Fast autoregressive transformers with linear
  attention.
\newblock In {\em International conference on machine learning}, pages
  5156--5165. PMLR, 2020.

\bibitem{khan2022transformers}
Salman Khan, Muzammal Naseer, Munawar Hayat, Syed~Waqas Zamir, Fahad~Shahbaz
  Khan, and Mubarak Shah.
\newblock Transformers in vision: A survey.
\newblock {\em ACM computing surveys (CSUR)}, 54(10s):1--41, 2022.

\bibitem{NEURIPS2022_8bb0d291}
Takeshi Kojima, Shixiang~(Shane) Gu, Machel Reid, Yutaka Matsuo, and Yusuke
  Iwasawa.
\newblock Large language models are zero-shot reasoners.
\newblock In {\em Advances in Neural Information Processing Systems},
  volume~35, pages 22199--22213. Curran Associates, Inc., 2022.

\bibitem{lee2022multi}
Kuang-Huei Lee, Ofir Nachum, Mengjiao~Sherry Yang, Lisa Lee, Daniel Freeman,
  Sergio Guadarrama, Ian Fischer, Winnie Xu, Eric Jang, Henryk Michalewski,
  et~al.
\newblock Multi-game decision transformers.
\newblock {\em Advances in Neural Information Processing Systems},
  35:27921--27936, 2022.

\bibitem{lin2010augmented}
Zhouchen Lin, Minming Chen, and Yi~Ma.
\newblock The augmented lagrange multiplier method for exact recovery of
  corrupted low-rank matrices.
\newblock {\em arXiv preprint arXiv:1009.5055}, 2010.

\bibitem{lin2017a}
Zhouhan Lin, Minwei Feng, Cicero~Nogueira dos Santos, Mo~Yu, Bing Xiang, Bowen
  Zhou, and Yoshua Bengio.
\newblock A structured self-attentive sentence embedding.
\newblock In {\em International Conference on Learning Representations}, 2017.

\bibitem{liu2021swin}
Ze~Liu, Yutong Lin, Yue Cao, Han Hu, Yixuan Wei, Zheng Zhang, Stephen Lin, and
  Baining Guo.
\newblock Swin transformer: Hierarchical vision transformer using shifted
  windows.
\newblock In {\em Proceedings of the IEEE/CVF International Conference on
  Computer Vision}, pages 10012--10022, 2021.

\bibitem{lu2019understanding}
Yiping Lu, Zhuohan Li, Di~He, Zhiqing Sun, Bin Dong, Tao Qin, Liwei Wang, and
  Tie-Yan Liu.
\newblock Understanding and improving transformer from a multi-particle dynamic
  system point of view.
\newblock {\em arXiv preprint arXiv:1906.02762}, 2019.

\bibitem{madry2018towards}
Aleksander Madry, Aleksandar Makelov, Ludwig Schmidt, Dimitris Tsipras, and
  Adrian Vladu.
\newblock Towards deep learning models resistant to adversarial attacks.
\newblock In {\em International Conference on Learning Representations}, 2018.

\bibitem{mao2022robust}
Xiaofeng Mao, Gege Qi, Yuefeng Chen, Xiaodan Li, Ranjie Duan, Shaokai Ye, Yuan
  He, and Hui Xue.
\newblock Towards robust vision transformer.
\newblock In {\em Proceedings of the IEEE/CVF conference on Computer Vision and
  Pattern Recognition}, pages 12042--12051, 2022.

\bibitem{DBLP:journals/corr/MerityXBS16}
Stephen Merity, Caiming Xiong, James Bradbury, and Richard Socher.
\newblock Pointer sentinel mixture models.
\newblock In {\em International Conference on Learning Representations}, 2017.

\bibitem{morris2020textattack}
John Morris, Eli Lifland, Jin~Yong Yoo, Jake Grigsby, Di~Jin, and Yanjun Qi.
\newblock Textattack: A framework for adversarial attacks, data augmentation,
  and adversarial training in nlp.
\newblock In {\em Proceedings of the 2020 Conference on Empirical Methods in
  Natural Language Processing: System Demonstrations}, pages 119--126, 2020.

\bibitem{nguyen2008robust}
Minh Nguyen and Fernando Torre.
\newblock Robust kernel principal component analysis.
\newblock {\em Advances in Neural Information Processing Systems}, 21, 2008.

\bibitem{nguyen2023mitigating}
Tam Nguyen, Tan Nguyen, and Richard Baraniuk.
\newblock Mitigating over-smoothing in transformers via regularized nonlocal
  functionals.
\newblock {\em Advances in Neural Information Processing Systems},
  36:80233--80256, 2023.

\bibitem{nguyen2024pidformer}
Tam Nguyen, C{\'e}sar~A Uribe, Tan~M Nguyen, and Richard~G Baraniuk.
\newblock Pidformer: Transformer meets control theory.
\newblock In {\em International Conference on Machine Learning}. PMLR, 2024.

\bibitem{nguyen2022improving}
Tam~Minh Nguyen, Tan~Minh Nguyen, Dung~DD Le, Duy~Khuong Nguyen, Viet-Anh Tran,
  Richard Baraniuk, Nhat Ho, and Stanley Osher.
\newblock Improving transformers with probabilistic attention keys.
\newblock In {\em International Conference on Machine Learning}, pages
  16595--16621. PMLR, 2022.

\bibitem{nguyen2022head}
Tan Nguyen, Tam Nguyen, Hai Do, Khai Nguyen, Vishwanath Saragadam, Minh Pham,
  Khuong~Duy Nguyen, Nhat Ho, and Stanley Osher.
\newblock Improving transformer with an admixture of attention heads.
\newblock {\em Advances in neural information processing systems},
  35:27937--27952, 2022.

\bibitem{nguyen2022fourierformer}
Tan Nguyen, Minh Pham, Tam Nguyen, Khai Nguyen, Stanley Osher, and Nhat Ho.
\newblock Fourierformer: Transformer meets generalized fourier integral
  theorem.
\newblock {\em Advances in Neural Information Processing Systems},
  35:29319--29335, 2022.

\bibitem{nguyen2021fmmformer}
Tan Nguyen, Vai Suliafu, Stanley Osher, Long Chen, and Bao Wang.
\newblock Fmmformer: Efficient and flexible transformer via decomposed
  near-field and far-field attention.
\newblock {\em Advances in neural information processing systems},
  34:29449--29463, 2021.

\bibitem{nguyen2023probabilistic}
Tan~M Nguyen, Tam Nguyen, Long Bui, Hai Do, Duy~Khuong Nguyen, Dung~D Le, Hung
  Tran-The, Nhat Ho, Stan~J Osher, and Richard~G Baraniuk.
\newblock A probabilistic framework for pruning transformers via a finite
  admixture of keys.
\newblock In {\em ICASSP 2023-2023 IEEE International Conference on Acoustics,
  Speech and Signal Processing (ICASSP)}, pages 1--5. IEEE, 2023.

\bibitem{nguyen2023a}
Tan~Minh Nguyen, Tam~Minh Nguyen, Nhat Ho, Andrea~L. Bertozzi, Richard
  Baraniuk, and Stanley Osher.
\newblock A primal-dual framework for transformers and neural networks.
\newblock In {\em The Eleventh International Conference on Learning
  Representations}, 2023.

\bibitem{nielsen2024elliptical}
Stefan~K Nielsen, Laziz~U Abdullaev, Rachel Teo, and Tan~M Nguyen.
\newblock Elliptical attention.
\newblock {\em Advances in Neural Information Processing Systems}, 2024.

\bibitem{papernot2018cleverhans}
Nicolas Papernot, Fartash Faghri, Nicholas Carlini, Ian Goodfellow, Reuben
  Feinman, Alexey Kurakin, Cihang Xie, Yash Sharma, Tom Brown, Aurko Roy,
  Alexander Matyasko, Vahid Behzadan, Karen Hambardzumyan, Zhishuai Zhang,
  Yi-Lin Juang, Zhi Li, Ryan Sheatsley, Abhibhav Garg, Jonathan Uesato, Willi
  Gierke, Yinpeng Dong, David Berthelot, Paul Hendricks, Jonas Rauber, and
  Rujun Long.
\newblock Technical report on the cleverhans v2.1.0 adversarial examples
  library.
\newblock {\em arXiv preprint arXiv:1610.00768}, 2018.

\bibitem{parikh-etal-2016-decomposable}
Ankur Parikh, Oscar T{\"a}ckstr{\"o}m, Dipanjan Das, and Jakob Uszkoreit.
\newblock A decomposable attention model for natural language inference.
\newblock In Jian Su, Kevin Duh, and Xavier Carreras, editors, {\em Proceedings
  of the 2016 Conference on Empirical Methods in Natural Language Processing},
  pages 2249--2255, Austin, Texas, November 2016. Association for Computational
  Linguistics.

\bibitem{DBLP:journals/corr/abs-2105-07581}
Sayak Paul and Pin-Yu Chen.
\newblock Vision transformers are robust learners.
\newblock In {\em Proceedings of the AAAI conference on Artificial
  Intelligence}, volume~36, pages 2071--2081, 2022.

\bibitem{pearson1901liii}
Karl Pearson.
\newblock Liii. on lines and planes of closest fit to systems of points in
  space.
\newblock {\em The London, Edinburgh, and Dublin philosophical magazine and
  journal of science}, 2(11):559--572, 1901.

\bibitem{radford2021learning}
Alec Radford, Jong~Wook Kim, Chris Hallacy, Aditya Ramesh, Gabriel Goh,
  Sandhini Agarwal, Girish Sastry, Amanda Askell, Pamela Mishkin, Jack Clark,
  et~al.
\newblock Learning transferable visual models from natural language
  supervision.
\newblock In {\em International Conference on Machine Learning}, pages
  8748--8763. PMLR, 2021.

\bibitem{radford2018improving}
Alec Radford, Karthik Narasimhan, Tim Salimans, and Ilya Sutskever.
\newblock Improving language understanding by generative pre-training.
\newblock {\em OpenAI report}, 2018.

\bibitem{radford2019language}
Alec Radford, Jeffrey Wu, Rewon Child, David Luan, Dario Amodei, and Ilya
  Sutskever.
\newblock Language models are unsupervised multitask learners.
\newblock {\em OpenAI blog}, 1(8):9, 2019.

\bibitem{JMLR:v21:20-074}
Colin Raffel, Noam Shazeer, Adam Roberts, Katherine Lee, Sharan Narang, Michael
  Matena, Yanqi Zhou, Wei Li, and Peter~J. Liu.
\newblock Exploring the limits of transfer learning with a unified text-to-text
  transformer.
\newblock {\em Journal of Machine Learning Research}, 21(140):1--67, 2020.

\bibitem{pmlr-v139-rao21a}
Roshan~M Rao, Jason Liu, Robert Verkuil, Joshua Meier, John Canny, Pieter
  Abbeel, Tom Sercu, and Alexander Rives.
\newblock Msa transformer.
\newblock In {\em Proceedings of the 38th International Conference on Machine
  Learning}, volume 139 of {\em Proceedings of Machine Learning Research},
  pages 8844--8856. PMLR, 18--24 Jul 2021.

\bibitem{rives2021biological}
Alexander Rives, Joshua Meier, Tom Sercu, Siddharth Goyal, Zeming Lin, Jason
  Liu, Demi Guo, Myle Ott, C~Lawrence Zitnick, Jerry Ma, et~al.
\newblock Biological structure and function emerge from scaling unsupervised
  learning to 250 million protein sequences.
\newblock {\em Proceedings of the National Academy of Sciences}, 118(15), 2021.

\bibitem{DBLP:journals/corr/abs-2110-11773}
Michael~E Sander, Pierre Ablin, Mathieu Blondel, and Gabriel Peyr{\'e}.
\newblock Sinkformers: Transformers with doubly stochastic attention.
\newblock In {\em International Conference on Artificial Intelligence and
  Statistics}, pages 3515--3530. PMLR, 2022.

\bibitem{schlag2021linear}
Imanol Schlag, Kazuki Irie, and J{\"u}rgen Schmidhuber.
\newblock Linear transformers are secretly fast weight programmers.
\newblock In {\em International Conference on Machine Learning}, pages
  9355--9366. PMLR, 2021.

\bibitem{shi2022revisiting}
Han Shi, Jiahui Gao, Hang Xu, Xiaodan Liang, Zhenguo Li, Lingpeng Kong, Stephen
  M.~S. Lee, and James Kwok.
\newblock Revisiting over-smoothing in {BERT} from the perspective of graph.
\newblock In {\em International Conference on Learning Representations}, 2022.

\bibitem{strudel2021segmenter}
Robin Strudel, Ricardo Garcia, Ivan Laptev, and Cordelia Schmid.
\newblock Segmenter: Transformer for semantic segmentation.
\newblock In {\em Proceedings of the IEEE/CVF international conference on
  computer vision}, pages 7262--7272, 2021.

\bibitem{subramanya2022backdoor}
Akshayvarun Subramanya, Aniruddha Saha, Soroush~Abbasi Koohpayegani, Ajinkya
  Tejankar, and Hamed Pirsiavash.
\newblock Backdoor attacks on vision transformers.
\newblock {\em arXiv preprint arXiv:2206.08477}, 2022.

\bibitem{1189643}
J.A.K. Suykens, T.~Van~Gestel, J.~Vandewalle, and B.~De~Moor.
\newblock A support vector machine formulation to pca analysis and its kernel
  version.
\newblock {\em IEEE Transactions on Neural Networks}, 14(2):447--450, 2003.

\bibitem{tang2021probabilistic}
Binh Tang and David~S. Matteson.
\newblock Probabilistic transformer for time series analysis.
\newblock In A.~Beygelzimer, Y.~Dauphin, P.~Liang, and J.~Wortman Vaughan,
  editors, {\em Advances in Neural Information Processing Systems}, 2021.

\bibitem{tarzanagh2023transformers}
Davoud~Ataee Tarzanagh, Yingcong Li, Christos Thrampoulidis, and Samet Oymak.
\newblock Transformers as support vector machines.
\newblock {\em arXiv preprint arXiv:2308.16898}, 2023.

\bibitem{tenney-etal-2019-bert}
Ian Tenney, Dipanjan Das, and Ellie Pavlick.
\newblock {BERT} rediscovers the classical {NLP} pipeline.
\newblock In {\em Proceedings of the 57th Annual Meeting of the Association for
  Computational Linguistics}, pages 4593--4601, Florence, Italy, July 2019.
  Association for Computational Linguistics.

\bibitem{DBLP:journals/corr/abs-2012-12877}
Hugo Touvron, Matthieu Cord, Matthijs Douze, Francisco Massa, Alexandre
  Sablayrolles, and Herv{\'e} J{\'e}gou.
\newblock Training data-efficient image transformers \& distillation through
  attention.
\newblock In {\em International conference on machine learning}, pages
  10347--10357. PMLR, 2021.

\bibitem{DBLP:journals/corr/abs-1904-13000}
Florian Tramer and Dan Boneh.
\newblock Adversarial training and robustness for multiple perturbations.
\newblock {\em Advances in neural information processing systems}, 32, 2019.

\bibitem{tran2024equivariant}
Viet-Hoang Tran, Thieu~N Vo, An~Nguyen The, Tho~Tran Huu, Minh-Khoi
  Nguyen-Nhat, Thanh Tran, Duy-Tung Pham, and Tan~Minh Nguyen.
\newblock Equivariant neural functional networks for transformers.
\newblock {\em arXiv preprint arXiv:2410.04209}, 2024.

\bibitem{tsai-etal-2019-transformer}
Yao-Hung~Hubert Tsai, Shaojie Bai, Makoto Yamada, Louis-Philippe Morency, and
  Ruslan Salakhutdinov.
\newblock Transformer dissection: An unified understanding for transformer{'}s
  attention via the lens of kernel.
\newblock In Kentaro Inui, Jing Jiang, Vincent Ng, and Xiaojun Wan, editors,
  {\em Proceedings of the 2019 Conference on Empirical Methods in Natural
  Language Processing and the 9th International Joint Conference on Natural
  Language Processing (EMNLP-IJCNLP)}, pages 4344--4353, Hong Kong, China,
  November 2019. Association for Computational Linguistics.

\bibitem{uesato2018adversarial}
Jonathan Uesato, Brendan O’donoghue, Pushmeet Kohli, and Aaron Oord.
\newblock Adversarial risk and the dangers of evaluating against weak attacks.
\newblock In {\em International Conference on Machine Learning}, pages
  5025--5034. PMLR, 2018.

\bibitem{vaswani2017attention}
Ashish Vaswani, Noam Shazeer, Niki Parmar, Jakob Uszkoreit, Llion Jones,
  Aidan~N Gomez, {\L}ukasz Kaiser, and Illia Polosukhin.
\newblock Attention is all you need.
\newblock In {\em Advances in neural information processing systems}, pages
  5998--6008, 2017.

\bibitem{vig-belinkov-2019-analyzing}
Jesse Vig and Yonatan Belinkov.
\newblock Analyzing the structure of attention in a transformer language model.
\newblock In {\em Proceedings of the 2019 ACL Workshop BlackboxNLP: Analyzing
  and Interpreting Neural Networks for NLP}, pages 63--76, Florence, Italy,
  August 2019. Association for Computational Linguistics.

\bibitem{voita-etal-2019-analyzing}
Elena Voita, David Talbot, Fedor Moiseev, Rico Sennrich, and Ivan Titov.
\newblock Analyzing multi-head self-attention: Specialized heads do the heavy
  lifting, the rest can be pruned.
\newblock In {\em Proceedings of the 57th Annual Meeting of the Association for
  Computational Linguistics}, pages 5797--5808, Florence, Italy, July 2019.
  Association for Computational Linguistics.

\bibitem{wang2022antioversmooth}
Peihao Wang, Wenqing Zheng, Tianlong Chen, and Zhangyang Wang.
\newblock Anti-oversmoothing in deep vision transformers via the fourier domain
  analysis: From theory to practice.
\newblock In {\em International Conference on Learning Representations}, 2022.

\bibitem{DBLP:journals/corr/abs-2006-04768}
Sinong Wang, Belinda~Z Li, Madian Khabsa, Han Fang, and Hao Ma.
\newblock Linformer: Self-attention with linear complexity.
\newblock {\em arXiv preprint arXiv:2006.04768}, 2020.

\bibitem{wang2022transtab}
Zifeng Wang and Jimeng Sun.
\newblock Transtab: Learning transferable tabular transformers across tables.
\newblock {\em Advances in Neural Information Processing Systems},
  35:2902--2915, 2022.

\bibitem{yang2021graph}
Yongyi Yang, Tang Liu, Yangkun Wang, Jinjing Zhou, Quan Gan, Zhewei Wei, Zheng
  Zhang, Zengfeng Huang, and David Wipf.
\newblock Graph neural networks inspired by classical iterative algorithms.
\newblock In {\em International Conference on Machine Learning}, pages
  11773--11783. PMLR, 2021.

\bibitem{NEURIPS2019_dc6a7e65}
Zhilin Yang, Zihang Dai, Yiming Yang, Jaime Carbonell, Russ~R Salakhutdinov,
  and Quoc~V Le.
\newblock Xlnet: Generalized autoregressive pretraining for language
  understanding.
\newblock In H.~Wallach, H.~Larochelle, A.~Beygelzimer, F.~d\textquotesingle
  Alch\'{e}-Buc, E.~Fox, and R.~Garnett, editors, {\em Advances in Neural
  Information Processing Systems}, volume~32. Curran Associates, Inc., 2019.

\bibitem{yuan2009sparse}
Xiaoming Yuan and Junfeng Yang.
\newblock Sparse and low-rank matrix decomposition via alternating direction
  method.
\newblock {\em Pacific Journal of Optimization}, 9:167, 2013.

\bibitem{zhang-feng-2021-modeling-concentrated}
Shaolei Zhang and Yang Feng.
\newblock Modeling concentrated cross-attention for neural machine translation
  with {G}aussian mixture model.
\newblock In Marie-Francine Moens, Xuanjing Huang, Lucia Specia, and Scott
  Wen-tau Yih, editors, {\em Findings of the Association for Computational
  Linguistics: EMNLP 2021}, pages 1401--1411, Punta Cana, Dominican Republic,
  November 2021. Association for Computational Linguistics.

\bibitem{zhang2019deep}
Shuai Zhang, Lina Yao, Aixin Sun, and Yi~Tay.
\newblock Deep learning based recommender system: A survey and new
  perspectives.
\newblock {\em ACM Computing Surveys (CSUR)}, 52(1):1--38, 2019.

\bibitem{zheng2022online}
Qinqing Zheng, Amy Zhang, and Aditya Grover.
\newblock Online decision transformer.
\newblock In Kamalika Chaudhuri, Stefanie Jegelka, Le~Song, Csaba Szepesvari,
  Gang Niu, and Sivan Sabato, editors, {\em Proceedings of the 39th
  International Conference on Machine Learning}, volume 162 of {\em Proceedings
  of Machine Learning Research}, pages 27042--27059. PMLR, 17--23 Jul 2022.

\bibitem{8100027}
Bolei Zhou, Hang Zhao, Xavier Puig, Sanja Fidler, Adela Barriuso, and Antonio
  Torralba.
\newblock Scene parsing through ade20k dataset.
\newblock In {\em 2017 IEEE Conference on Computer Vision and Pattern
  Recognition (CVPR)}, pages 5122--5130, 2017.

\bibitem{zhou2022understanding}
Daquan Zhou, Zhiding Yu, Enze Xie, Chaowei Xiao, Animashree Anandkumar, Jiashi
  Feng, and Jose~M Alvarez.
\newblock Understanding the robustness in vision transformers.
\newblock In {\em International Conference on Machine Learning}, pages
  27378--27394. PMLR, 2022.

\bibitem{zhuang-etal-2021-robustly}
Liu Zhuang, Lin Wayne, Shi Ya, and Zhao Jun.
\newblock A robustly optimized {BERT} pre-training approach with post-training.
\newblock In {\em Proceedings of the 20th Chinese National Conference on
  Computational Linguistics}, pages 1218--1227, Huhhot, China, August 2021.
  Chinese Information Processing Society of China.

\end{thebibliography}
\bibliographystyle{plain}

\newpage 
\appendix

\begin{center}
{\bf \Large{Supplement to ``Unveiling the Hidden Structure of Self-Attention
via Kernel Principal Component Analysis''}}
\end{center}
\DoToC

\section{Experiment Details}
\label{appx:exp_details}
\paragraph{Implementation Details of RPC-SymViT:} Our RPC-SymViT models have 5.2M parameters, the same as the SymViT baseline. We use a standard tiny configuration with 12 transformer layers, 3 attention heads per layer, and a model dimension of 192 and simply replace softmax attention with RPC-Attention. We follow the training settings as in \cite{DBLP:journals/corr/abs-2012-12877} and their implementation is available at \url{https://github.com/facebookresearch/deit}. In a Segmenter model, we use the same RPC-SymViT setting to replace the baseline SymViT backbone. We follow the training details in \cite{strudel2021segmenter} and their code is publicly available as well, \url{https://github.com/rstrudel/segmenter}. 

There are 3 hyperparameters: 1) $\mu$: this parameter controls the singular value thresholding operator in the PAP algorithm. We set $\mu$ to the recommended value given in Definition~\ref{def:rpc-attn}; 2) $\lambda$: this is a regularization parameter that controls the sparsity of the corruption matrix $\mS$. We finetune $\lambda$ for training and observe that RPC-SymViT with $\lambda=3$ yields the best performance for models with 2 iterations per layer and $\lambda=4$ yields the best performance for models with iterations only in the first layer; 3) $n$: the number of iterations of the PAP algorithm in a RPC-Attention layer. 

\paragraph{Implementation Details of RPC-Symlm:} For our language model, we use a standard transformer language model \cite{vaswani2017attention}, with a symmetric attention layer. The model has a dimension of 128 for the keys, queries and values, while the training and evaluation context length is set at 256. There are 16 layers altogether and 8 heads per layer. Similarly, we replace the softmax attention with RPC-Attention only in the first four layers to save on computational overhead. There are the same 3 hyperparameters as in RPC-SymViT and we use the same value for $\mu$, $\lambda=4$ and $n=4$. We also follow the standard training settings as in \cite{schlag2021linear,DBLP:journals/corr/MerityXBS16} and the code base developed by \cite{schlag2021linear}, available here \url{https://github.com/IDSIA/lmtool-fwp}. 

\subsection{Robustness against Data Corruptions}
\label{appx:robust_corrupt_details}
\paragraph{Datasets:} We use the ImageNet-1K dataset that contains 1.28M training images and 50K validation images. There are 1000 classes of images and the model learns an image classification task. For robustness to common corruptions, we use ImageNet-C (IN-C)~\cite{DBLP:journals/corr/abs-1903-12261} which consists of 15 different types of corruptions applied to the ImageNet-1K validation set with 5 levels of severity. To test robustness to both input data distribution shifts as well as label distribution shifts, we use ImageNet-A (IN-A) and ImageNet-O (IN-O)~\cite{hendrycks2021natural} respectively. Both of these datasets contain a 200 class subset of ImageNet-1K classes with adversarially filtered images. Finally, we test our model on ImageNet-R (IN-R)~\cite{DBLP:journals/corr/abs-2006-16241} which contains various artistic renditions of images. This evaluates the model's generalization ability to abstract visual renditions. 

\paragraph{Metrics:} On ImageNet-1K, ImageNet-A and ImageNet-R, we report the top-1 accuracies for all experiments. We include top-5 accuracies on ImageNet-1K. On ImageNet-C, the standard metric for evaluation is the mean Corruption Error (mCE). To calculate this, we average the top-1 error rate for each corruption type across the 5 levels of severity and divide them by AlexNet's average errors, then take the final average across all corruption types. We report the area under the precision-recall curve (AUPR) for ImageNet-O which requires anomaly scores. The score is obtained by taking the negative of the highest softmax probability output by the model. The direction of increasing or decreasing values of these metrics signifying greater robustness will be indicated in the table with an arrow. 

\subsection{Robustness under Adversarial Attacks}
\label{appx:attack}
\paragraph{Attacks:} We evaluate the robustness of our method against adversarial attacks using CleverHans \cite{papernot2018cleverhans} and AutoAttack \cite{croce2020reliable}. All attacks are executed on ImageNet-1K's validation set, and each model is evaluated on the whole set. In particular, we use untargeted, white box attacks such as PGD, FGSM, SLD and CW. In addition, we provide results on gradient-free black box attack, SPSA, diverse AutoAttack, in the standard setting and a simple noise-adding attack. AutoAttack consists of untargeted APGD-CE, targeted APGD-DLR, targeted Fast Adaptive Boundary (FAB) and Square Attack. For further justification of the benefit of our method, we evaluate a variant of our model, RPC-SymViT (6iter/layer1) solely on the black-box Square Attack and show that we are robust against black box attacks as well. This is provided in Appendix \ref{appx:square}. We use a perturbation budget of $\epsilon=1/255$ with the $l_\infty$ norm to manipulate the images and evaluate each model with an incremental increase in perturbation for PGD and FGSM. In SPSA, we run 40 steps per attack with a perturbation budget of $\epsilon=0.1$. PGD attack uses a step size of $\alpha=0.15$ for 20 steps, where at each step the image is adjusted slightly to maximize the model's loss. Similarly, FGSM does this for a single step. For each attack, we report the top-1 and top-5 accuracy of the model on the distorted dataset with the maximum perturbation budget in Table~\ref{tab:table3} and across all different perturbations in Figure~\ref{fig:attack}.

\subsection{ADE20K Image Segmentation}
\label{app:ade20k}
\paragraph{Dataset:} The ADE20K dataset includes complex scenes featuring highly detailed labels, making it one of the most challenging segmentation tasks. The training set contains 20,210 images spread across 150 distinct semantic categories. The validation set includes 2,000 images, while the test set comprises 3,352 images. The metrics report for this task are the Mean Accuracy (\%) and Mean Intersection-Over-Union (IOU).  

\subsection{WikiText-103 Language Modeling}
\paragraph{Dataset:} The WikiText-103 dataset ~\cite{DBLP:journals/corr/MerityXBS16} is derived from Wikipedia articles and is designed to capture long-range contextual dependencies. The training set contains about 28,000 articles, with a total of 103 million words. Each article is divided into text blocks with approximately 3,600 words. The validation and test sets have 218,000 and 246,000 words, respectively, with both sets comprising 60 articles and totaling about 268,000 words. For evaluation, we use a batch size of $1$ and apply a sliding window of length $L$ to process the text sequences. When calculating perplexity (PPL), we focus only on the final position, except in the first segment, where we evaluate all positions. We corrupt the both validation and test datasets to demonstrate the
 robustness of RPC-Attention using TextAttack's word swap attack \cite{morris2020textattack} to create the attacked WikiText-103 dataset. This adversarial attack randomly replaces words in the dataset with a generic ``AAA'' for evaluation making it difficult for the model to predict the next word in the sequence correctly. 

\section{Calculating the Gram Matrix \texorpdfstring{$\widetilde{\mK}_{\vvarphi}$}{}}
\label{appx:comp_gram}
In this section, we will show that the centered Gram matrix $\widetilde{\mK}_{\vvarphi}$ can be computed from the uncentered Gram matrix ${\mK}_{\vvarphi}$ with elements $\mK_{\vvarphi}(i,j) = k_{\vvarphi}(\vk_i, \vk_j) = \vvarphi(\vk_i)^{\top}\vvarphi(\vk_j)$. In particular, $\widetilde{\mK}_{\vvarphi} = {\mK}_{\vvarphi} - \textbf{1}_{N}{\mK}_{\vvarphi} - {\mK}_{\vvarphi}\textbf{1}_{N} + \textbf{1}_{N}{\mK}_{\vvarphi}\textbf{1}_{N}$, where $\textbf{1}_{N}$ denotes the $N \times N$ matrix in which every element takes the value $1/N$. 
Our centered feature vector has the form
\begin{align*}
    \tilde{\vvarphi}(\vk_j) =  \vvarphi(\vk_j) - \frac{1}{N}\sum_{j'=1}^{N}\vvarphi(\vk_{j'}).
\end{align*} Then the elements of the centered Gram matrix are given as follows
\begin{align*}
    \widetilde{\mK}_{\vvarphi}(i,j) &= \tilde{k}_{\vvarphi}(\vk_i, \vk_j) \\
    &= \tilde{\vvarphi}(\vk_i)^\top \tilde{\vvarphi}(\vk_j) \\
    &= \vvarphi(\vk_i)^\top \vvarphi(\vk_j) - \frac{1}{N}\sum_{j'=1}^{N}\vvarphi(\vk_i)^\top \vvarphi(\vk_{j'}) - \frac{1}{N}\sum_{j'=1}^{N}\vvarphi(\vk_{j'})^\top \vvarphi(\vk_{j}) + \frac{1}{N^2}\sum_{j'=1}^{N}\sum_{l=1}^{N}\vvarphi(\vk_{j'})^\top \vvarphi(\vk_{l}) \\
    &= k_{\vvarphi}(\vk_i, \vk_j) -  \frac{1}{N}\sum_{j'=1}^{N} k_{\vvarphi}(\vk_i, \vk_{j'}) -  \frac{1}{N}\sum_{j'=1}^{N} k_{\vvarphi}(\vk_{j'}, \vk_{j}) + \frac{1}{N^2}\sum_{j'=1}^{N}\sum_{l=1}^{N}k_{\vvarphi}(\vk_{j'}, \vk_{l})
\end{align*}
which expressed in matrix form, gives the result. 

\section{Plotting $J_{\text{proj}}$ in Section~\ref{subsubsec:J_proj}}
\label{appx:J_proj}
In Section~\ref{subsubsec:J_proj}, we plotted the reconstruction loss at each epoch of training. Here, we provide the details of the calculation of this loss. 
From Eqn.~(\ref{eqn:proj-cost}),
\begin{align*}
    L &= \frac{1}{N}\sum_{i=1}^{N}\|\vvarphi(\vq_i)-\sum_{d=1}^{D_v}h_{id}u_d\|^2 \\
    &= \frac{1}{N}\sum_{i=1}^{N}(\|\vvarphi(\vq_i)\|^2 - \|\vh_i\|^2). 
\end{align*}
In the above, $\vh_i$ is simply our attention output and $\|\vvarphi(\vq_i)\|^2 = \vvarphi(\vq_i)^\top\vvarphi(\vq_i) = \frac{e^{\vq_i^\top \vq_i/\sqrt{D}}}{(\sum_{j=1}^N e^{\vq_i^\top \vk_j}/\sqrt{D})^2}$ for $i=1,\hdots,N$, can be calculated using the kernel trick as follows. Let $\mA$ be the softmax attention matrix, $\va_1$ its first column and $\va_1^2$ denote the element-wise product. 
\begin{align*}
    log(\va_1^2) &= log(
    \begin{bmatrix}
        \frac{e^{2 \vq_1^\top \vk_1/\sqrt{D}}}{(\sum_{j=1}^N e^{\vq_1^\top \vk_j/\sqrt{D}})^2} \\
        \vdots \\
        \frac{e^{2 \vq_N^\top \vk_1/\sqrt{D}}}{(\sum_{j=1}^N e^{\vq_N^\top \vk_j/\sqrt{D}})^2}
    \end{bmatrix}) \\
    &= log(
    \begin{bmatrix}
        e^{2 \vq_1^\top \vk_1/\sqrt{D}} \\
        \vdots \\
        e^{2 \vq_N^\top \vk_1/\sqrt{D}}
    \end{bmatrix}) - log(
    \begin{bmatrix}
        (\sum_{j=1}^N e^{\vq_1^\top \vk_j/\sqrt{D}})^2 \\
        \vdots \\
        (\sum_{j=1}^N e^{\vq_N^\top \vk_j/\sqrt{D}})^2
    \end{bmatrix})
\end{align*}
\begin{align*}
\implies log(
    \begin{bmatrix}
        (\sum_{j=1}^N e^{\vq_1^\top \vk_j/\sqrt{D}})^2 \\
        \vdots \\
        (\sum_{j=1}^N e^{\vq_N^\top \vk_j/\sqrt{D}})^2
    \end{bmatrix}) &=  log(
    \begin{bmatrix}
        e^{2 \vq_1^\top \vk_1/\sqrt{D}} \\
        \vdots \\
        e^{2 \vq_N^\top \vk_1/\sqrt{D}}
    \end{bmatrix})- log(\va_1^2) \\
    &= \begin{bmatrix}
        2 \vq_1^\top \vk_1/\sqrt{D} \\
        \vdots \\
        2 \vq_N^\top \vk_1/\sqrt{D}
    \end{bmatrix}- log(\va_1^2)
\end{align*}
\begin{align*}
\implies 
    \begin{bmatrix}
        (\sum_{j=1}^N e^{\vq_1^\top \vk_j/\sqrt{D}})^2 \\
        \vdots \\
        (\sum_{j=1}^N e^{\vq_N^\top \vk_j/\sqrt{D}})^2
    \end{bmatrix} = e^{
    \begin{bmatrix}
        2 \vq_1^\top \vk_1/\sqrt{D} \\
        \vdots \\
        2 \vq_N^\top \vk_1/\sqrt{D}
    \end{bmatrix} - log(\va_1^2)}
\end{align*}
Then,
\begin{align*}
    \begin{bmatrix}
        \|\vvarphi(\vq_1)\|^2 \\
        \vdots \\
        \|\vvarphi(\vq_N)\|^2
    \end{bmatrix} = \begin{bmatrix}
        e^{\vq_1^\top \vq_1/\sqrt{D}} \\
        \vdots \\
        e^{\vq_N^\top \vq_N/\sqrt{D}}
    \end{bmatrix} / e^{
    \begin{bmatrix}
        2 \vq_1^\top \vk_1/\sqrt{D} \\
        \vdots \\
        2 \vq_N^\top \vk_1/\sqrt{D}
    \end{bmatrix} - log(\va_1^2)}
\end{align*}

\section{Principal Component Pursuit}
\label{appx:pcp}
Let $\mM, \mL, \mS \in \RR^{N\times D}$ be the matrix of our corrupted measurements, the low-rank matrix we seek to recover and a sparse corruption matrix respectively. The optimization problem we would like to solve is 
\begin{equation}
\label{eq:sdp-appx}
  \begin{array}{ll}
    \text{minimize}_{\mL, \mS}   & \quad \|\mL\|_* + \lambda \|\mS\|_1\\
    \text{subject to} & \quad \mL + \mS = \mM 
  \end{array}
\end{equation}
Under minimal assumptions that the low-rank component $\mL$ is not sparse (i.e., $\mL$ satisfies the incoherence condition defined in~\cite{candes2009robust}), and the sparse component $\mS$ is not low-rank (i.e., the sparsity pattern of $S$ is selected uniformly at random), we will follow the author's choice of algorithm to solve the PCP which is to use the Alternating Direction Method of Multipliers (ADMM). 

The ADMM algorithm uses the augmented Lagrangian, 
\begin{align*}
l(\mL,\mS,\mY)=\|\mL\|_*+\lambda \|\mS\|_1+\langle 
    \mY,\mM-\mL-\mS \rangle + \mu/2\|\mM-\mL-\mS\|_F^2
\end{align*}
and solves a sequence of optimization problems. We iterate through setting $\mS_{k+1}=\arg\min_\mS l(\mL_k,\mS,\mY_k)$ and $\mL_{k+1}=\arg\min_\mL l(\mL,\mS_{k+1},\mY_k)$ before updating the Langrange multiplier matrix $\mY_{k+1}=\mY_k+\mu (\mM-\mL_{k+1}-\mS_{k+1})$. The advantage of the algorithm is that it turns a complicated optimisation problem into sub problems that have straightforward and efficient solutions. Without much difficulty we can show that:
\begin{align*}
    \arg\min_\mS l(\mL,\mS,\mY) =& \mathcal{\mS_{\lambda / \mu}}(\mM-\mL+\mu^{-1}\mY)\\
    \arg\min_\mL l(\mL,\mS,\mY) =& \mathcal{D_\mu}(\mM-\mS-\mu^{-1}\mY)
\end{align*}
where $\mathcal{S_\tau}(x)=\text{sgn}(x)\max(|x|-\tau,0)$ is the shrinkage operator, extended to matrices by applying it element-wise and $\mathcal{D_\tau}(\mX)=\mU\mathcal{S_\tau}(\Sigma)\mV^*$ is the singular value thresholding operator with the singular value decomposition of $\mX=\mU \Sigma \mV^*$. 

The algorithm is summarised as below
\begin{algorithm}[H]
\begin{algorithmic}
{\small
    \STATE \textbf{initialize:} $\mS_0 = \mY_0 = \textbf{0}$; $\mu, \lambda > 0$.
    \WHILE{not converged}
      \STATE compute $\mS_{k+1} = \mathcal{\mS_{\lambda/\mu}}(\mM - \mL_k + \mu^{-1}\mY_k)$;
      \STATE compute $\mL_{k+1} = \mathcal{D_\mu}(\mM - \mS_{k+1} - \mu^{-1}\mY_k)$;
      \STATE compute $\mY_{k+1} = \mY_k + \mu(\mM - \mL_{k+1} - \mS_{k+1})$;
    \ENDWHILE
    \STATE \textbf{output:} $\mL$, $\mS$.}
  \end{algorithmic}
 \caption{\small ADMM for Principal Components Pursuit}
\end{algorithm}

\section{Additional Experimental Results}
\label{appx:extra_experiments}

\subsection{Reparameterization of Self-Attention}
\label{appx:reparam}
\begin{figure}[!t]
  \centering  \includegraphics[width=0.8\linewidth]{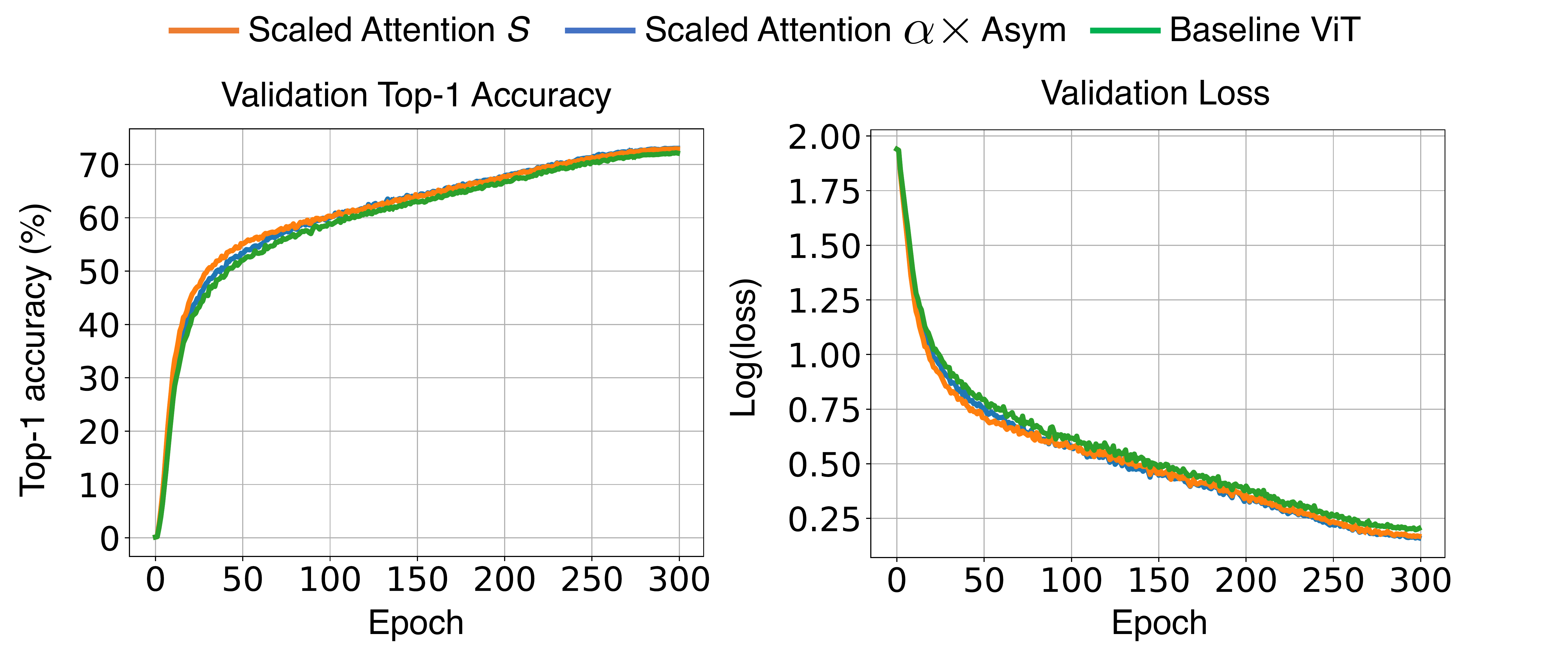}
  \caption{\small Plot of the validation top-1 accuracy (\%) and loss on a log scale of the baseline asymmetric attention ViT and two variants with the parameterization of Remark.~\ref{rm:parameterization}. The curves are plotted for the full training time and show $\mS$ trained as a matrix parameter as well as a scalar parameter scaling a symmetric attention matrix.}
  \label{fig:S_param_all}
\end{figure}

Let $\mA_{sym}=[a_{ij}]$ be the softmax attention matrix. Then, from Remark~\ref{rm:parameterization}, $\mS$ has the following form and we multiply the numerator and denominator by $1/K(\vk_j,\vk_{j'})$ to obtain a much more convenient expression. 
\begin{align*}
    s_{j'j}= &\frac{g(\vk_j)}{g(\vk_{j'})} \\
    = &\frac{g(\vk_j)}{g(\vk_{j'})} \times \frac{1/K(\vk_j,\vk_{j'})}{1/K(\vk_j,k_{\vj'})} \\
    = &\frac{g(\vk_j)}{K(\vk_j,\vk_{j'})} \div \frac{g(\vk_{j'})}{K(\vk_j,\vk_{j'})} \\
    = &\frac{g(\vk_j)}{K(\vk_j,\vk_{j'})} \times \frac{K(\vk_j,\vk_{j'})}{g(\vk_{j'})} \\
    = &\frac{a_{j'j}}{a_{jj'}} \\
    \implies \mS = &\frac{1}{N}\mA_{sym} \odot 1/\mA_{sym}^\top
\end{align*}

In Fig. \ref{fig:S_param_all}, we plot the full training curve of the two versions of Scaled Attention, Scaled Attention $S$ and Scaled Attention $\alpha \times \text{Asym}$. 
We observe that the parameterized models with Scaled Attention do converge more quickly and even obtain a slightly higher validation top-1 accuracy of 73.02\% for the scalar variant as compared to the standard asymmetric ViT at 72.18\%. The final validation loss is also lower at 1.18 and 1.23 respectively. 

\subsection{Pairwise Absolute Differences of $\gamma_i$ and $\gamma_j$}
\label{appx:eigdiff}
\begin{figure*}[t!]
  \centering  \includegraphics[width=1.0\linewidth]{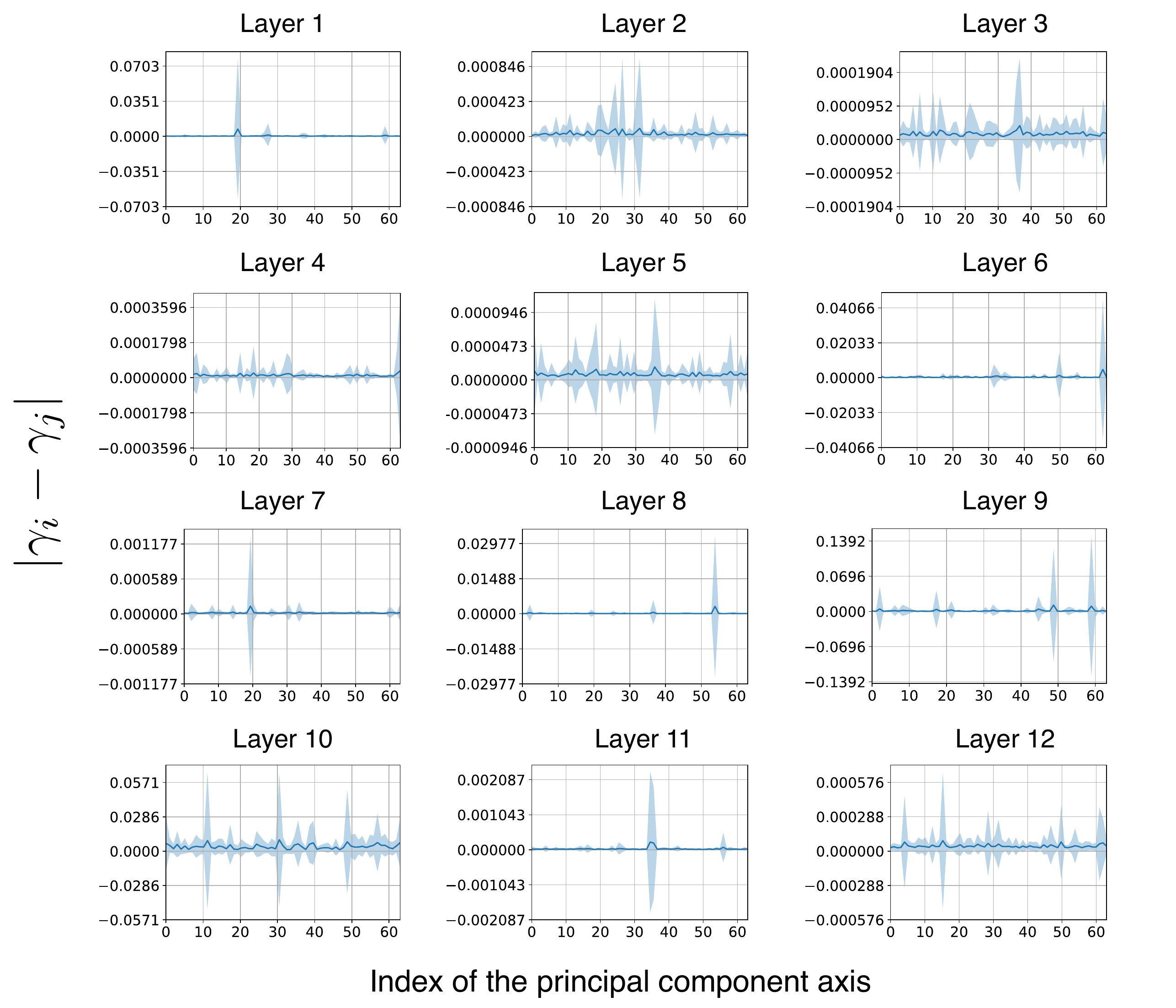}
  \caption{\small Plot of the mean and standard deviation of the differences in coordinate values of constant vector $\mathbf{1}\lambda_d$ for $d=1,\dots,D_v$ for all 12 layers of a ViT-tiny model. The mean should be $0$ with small standard deviations when $v_{dj}\approx \frac{a_{dj}}{g(\vk_j)} - \frac{1}{N}\sum_{j'=1}^N\frac{a_{dj'}}{g(\vk_{j})}$.}
  \label{fig:eigenval_diffall}
\end{figure*}

In Figure~\ref{fig:eigenval_diffall}, we provide the plot of the absolute differences of the coordinates of $\frac{\widetilde{\mK}_{\vvarphi}\hat{\va}_d}{N\hat{\va}_d}$ in all 12 layers of a ViT-tiny model as elaborated in Section~\ref{subsubsec:learning_eg}. This would be a constant vector within our framework and the plots provide empirical evidence to justify our claim. In all of the layers, the means are noticeably close to 0 and the standard deviations are also small suggesting that we indeed recovered a constant vector. 

\subsection{ImageNet-C Results by Corruption Type}
\label{appx:allimC}
In Table~\ref{tab:table6}, we provide the full results of various RPC-SymViT models against the standard SymViT on ImageNet-C for all corruption types averaged over the 5 severity levels. In all types, except for Zoom Blur, the RPC-SymViT model with 6 iterations in the 1st layer outperforms SymViT. 

\begin{table*}[!t]
{\footnotesize	
\setlength{\tabcolsep}{4.7 pt}
  \begin{center}
    \caption{\small Top-1 accuracy (\%) and mean corruption error (mCE) of all RPC-SymViT variants and SymViT on each corruption type in ImageNet-C. RPC-SymViT ($n$iter/layer1) applies $n$ PAP iterations only at the first layer. RPC-SymViT ($n$iter/all-layer) applies $n$ PAP iterations at all layers. }
    \label{tab:table6}
    \begin{tabular}{llccccc} 
    \toprule
      \multirow{2}{*}{Corruption Type} & Model/ & SymViT & RPC-SymViT & RPC-SymViT & RPC-SymViT &  RPC-SymViT \\
      & Metric & & (2iter/all-layer) & (4iter/layer1) & (5iter/layer1) & (6iter/layer1) \\
        \hline
      \multirow{2}{*}{Brightness} & Top-1 $\uparrow$ & 63.21  & 62.97  & 63.74  & 64.19 & 64.31  \\
      & mCE $\downarrow$ & 65.16 & 65.69 & 64.22 & 63.43 & 63.22
      \\ 
      \hline
      \multirow{2}{*}{Contrast} & Top-1 $\uparrow$ & 50.25 & 49.59 & 49.30 & 49.84  & 50.18  \\
      & mCE $\downarrow$& 58.53  & 59.08& 59.42 & 58.79& 58.39 
      \\
      \hline
      \multirow{2}{*}{Defocus Blur} & Top-1 $\uparrow$ & 35.47  & 35.61 & 35.54  & 35.26  & 36.38  \\
      & mCE $\downarrow$& 78.71& 78.53 & 78.62& 78.96& 77.59
      \\
      \hline
      \multirow{2}{*}{Elastic Transform}& Top-1 $\uparrow$ & 44.26 & 44.52& 44.68  & 44.84 & 44.72 \\
        & mCE $\downarrow$& 86.28  & 85.87 & 85.63& 85.38 & 85.56 
      \\
      \hline
      \multirow{2}{*}{Fog}& Top-1 $\uparrow$ & 42.72 & 45.49 & 46.03 & 46.59 & 46.40 \\
       & mCE $\downarrow$& 69.91 & 66.53 & 65.87 & 65.19 & 65.42 
      \\
      \hline
      \multirow{2}{*}{Frost}& Top-1 $\uparrow$ & 44.55 & 45.06 & 45.93 & 46.24 & 46.27 \\
      & mCE $\downarrow$& 67.08 & 66.46 & 65.41 & 65.04 & 65.01 
      \\
      \hline
      \multirow{2}{*}{Gaussian Noise} & Top-1 $\uparrow$ &40.97 & 41.64 & 42.22 & 41.67 & 42.51  \\
      & mCE $\downarrow$& 66.59 & 65.84& 65.19 & 65.80 & 64.85
      \\
      \hline
      \multirow{2}{*}{Glass Blur} & Top-1 $\uparrow$ & 27.79 & 28.12 & 28.48 & 27.87 & 28.05 \\
      & mCE $\downarrow$& 87.39 & 87.00 & 86.56 & 87.30 & 87.07 
      \\
      \hline
      \multirow{2}{*}{Impulse Noise} & Top-1 $\uparrow$ & 38.56 & 38.98 & 40.24 & 39.46 & 40.33 \\
      & mCE $\downarrow$& 66.59 & 66.14 & 64.77 & 65.61 & 64.67 
      \\
      \hline
      \multirow{2}{*}{JPEG Compression} & Top-1 $\uparrow$ & 46.28 & 46.23 & 47.92 & 48.41 & 48.63 \\
      & mCE $\downarrow$& 88.58 & 88.65& 85.87  & 85.06 & 84.70 
      \\
      \hline
      \multirow{2}{*}{Motion Blur}& Top-1 $\uparrow$  & 40.47 & 40.00 & 41.72 & 41.81 & 42.22\\
      & mCE $\downarrow$& 75.75 & 76.43 & 74.16 & 74.04  & 73.51 
      \\
      \hline
      \multirow{2}{*}{Pixelate}& Top-1 $\uparrow$  & 39.36 & 39.81 & 41.13 & 41.86 & 42.25 \\
      & mCE $\downarrow$& 84.47 & 83.84 & 82.01 & 81.00 & 80.45 
      \\
      \hline
      \multirow{2}{*}{Shot Noise}& Top-1 $\uparrow$  & 38.83 & 39.20 & 39.52 & 39.03 & 39.71 \\
      & mCE $\downarrow$& 68.38 & 67.98 & 67.61 & 68.16 & 67.40 
      \\
      \hline
      \multirow{2}{*}{Snow}& Top-1 $\uparrow$  & 38.27 & 37.87 & 38.75  & 39.12& 38.90 \\
      & mCE $\downarrow$& 71.22 & 71.67 & 70.66  & 70.23 & 70.48 
      \\
      \hline
      \multirow{2}{*}{Zoom Blur}& Top-1 $\uparrow$  & 30.69 & 29.49  & 29.97 & 30.20& 30.48 \\
      & mCE $\downarrow$& 86.82 & 88.32& 87.72  & 87.43 & 87.08 
      \\
      \bottomrule
    \end{tabular}
  \end{center}}
\end{table*}
\subsection{Square Attack}
\label{appx:square}
Square attack \cite{andriushchenko2020square} is a score-based black box attack that does not use local gradient information, hence it is not affected by gradient masking. It operates based on a random search scheme that modifies square-shaped regions such that at each iteration, the perturbation lies approximately at the boundary of the feasible set. We evaluate our RPC-SymViT(6iter/layer1) variant on square attacked ImageNet-1K validation set and compare it to the baseline. Our result of the top-1 and top-5 accuracy in Table \ref{tab:table8} illustrates that the effectiveness of RPC-Attention against adversarial attacks is not solely due to a form gradient masking as we still significantly outperform the baseline on square. 
\begin{table*}[t!]
{\footnotesize	
  \begin{center}
    \caption{\small Top-1 and Top-5 accuracy (\%) of RPC-SymViT(6iter/layer1) and baseline SymViT on square attacked ImageNet-1K validation data. RPC-SymViT ($n$iter/layer1) applies $n$ PAP iterations only at the first layer. }
    \label{tab:table8}
    \begin{tabular}{lcc} 
    \toprule
      \multirow{2}{*}{Model} & \multicolumn{2}{c}{Square attack}\\
      & Top-1 $\uparrow$ & Top-5 $\uparrow$ \\
        \midrule
      SymViT (baseline) & 41.50 & 79.79 \\
      \midrule
      RPC-SymViT (6iter/layer1) & \textbf{43.64} & \textbf{80.81} 
       \\
    \bottomrule
    \end{tabular}
  \end{center} }
\end{table*}
\subsection{Results on ADE20K Image Segmentation}
We further evaluate the benefits of our method by implementing RPC-Attention in a Segmenter model \cite{strudel2021segmenter} and providing results on the ADE20K image segmentation task \cite{8100027} in Table \ref{tab:table4}. We report the Mean Accuracy and Mean Intersection-Over-Union (IOU) for a  Segmenter with a RPC-SymViT backbone and compare it against a baseline SymViT backbone. To assess the robustness of each model, we corrupt the ADE20K dataset with the same 15 corruption types in ImageNet-C 
and report the same metrics averaged 
over all corruption types in Table \ref{tab:table4} as well. 
Details of our implementation can be found in Appendix \ref{app:ade20k}. On both datasets, RPC-SymViT outperforms the baseline substantially. 
\begin{table*}[!t]
{\footnotesize	
  \begin{center}
    \caption{\small Mean accuracy (\%) and mean Intersection-Over-Union (IOU) of RPC-SymViT and SymViT on clean ADE20K and corrupted ADE20K dataset. RPC-SymViT ($n$iter/layer1) applies $n$ PAP iterations only at the first layer. RPC-SymViT ($n$iter/all-layer) applies $n$ PAP iterations at all layers.}
    \label{tab:table4}
    \begin{tabular}{lcccc} 
    \toprule
      \multirow{2}{*}{Model} & \multicolumn{2}{c}{ADE20K}& \multicolumn{2}{c}{Corrupted ADE20K}  \\
      & Mean Acc. $\uparrow$ & Mean IOU $\uparrow$ & Mean Acc. $\uparrow$ & Mean IOU $\uparrow$ \\
        \hline
      SymViT &  44.27 & 34.00 & 14.85 & 10.47 \\
      \midrule
      RPC-SymViT (4iter/layer1)  & \textbf{45.61} & \textbf{34.69} & 16.06 & 11.47  \\
      RPC-SymViT (5iter/layer1)  & 45.51 & 34.63 & 16.15 & 11.35 \\
      RPC-SymViT (6iter/layer1)  & 45.27 & 34.24 & \textbf{16.44} & \textbf{11.60} \\
      \midrule
      RPC-SymViT (2iter/all-layer) &  43.61 & 33.5& 15.04 & 10.64 \\
    \bottomrule
    \end{tabular}
  \end{center}
  }
\end{table*}

\subsection{Results on RPC-SymViT-base}
To show that RPC-Attention is not limited to small-scale models, we conduct experiments on a larger model, SymViT-base with 12 transformer layers, 12 heads per layer, and a hidden dimension of 768. We train a SymViT-base with RPC-Attention in all layers using 2 iterations on ImageNet-1K for 300 epochs. We refer to this model as RPC-SymViT-base (2iter/all-layer). We compare our RPC-SymViT-base with the baseline on the same standard robustness benchmarks as before, ImageNet-C, ImageNet-A, and ImageNet-O, as well as on white box attacks PGD and FGSM with the highest perturbation budgets. These results are in Table \ref{tab:table9} and \ref{tab:table10} respectively and show that RPC-Attention is also effective in SymViT-base.
\begin{table*}[!t]
{ \footnotesize	
  \begin{center}
    \caption{\small Top-1, top-5 accuracy (\%) , mean corruption error (mCE), and area under the precision-recall curve (AUPR) of RPC-SymViT-base and SymViT-base on clean ImageNet-1K data and popular standard robustness benchmarks for image classification. RPC-SymViT-base ($n$iter/all-layer) applies $n$ PAP iterations at all layers. }
    \label{tab:table9}
   \begin{tabular}{lccccccc} 
    \toprule
      \multirow{2}{*}{Model} & \multicolumn{2}{c}{IN-1K} & IN-A & \multicolumn{2}{c}{IN-C} & IN-O \\
      & Top-1 $\uparrow$ & Top-5 $\uparrow$ & Top-1 $\uparrow$ & Top-1 $\uparrow$ & mCE $\downarrow$ & AUPR $\uparrow$\\
        \midrule
      SymViT-base (baseline) & 80.62 & 94.78 & 24.03 & 58.88 & 52.57 & 23.96 \\
      \midrule
      RPC-SymViT-base (2iter/all-layer) & \textbf{80.72} & \textbf{94.82} & \textbf{24.97} & \textbf{59.29} & \textbf{52.00} & \textbf{26.01} \\
    \bottomrule
    \end{tabular}
  \end{center}
}
\end{table*}
\begin{table*}[t!]
{\footnotesize	
  \begin{center}
    \caption{\small Top-1 and top-5 accuracy (\%) of RPC-SymViT-base and SymViT-base on PGD and FGSM attacked ImageNet-1K validation data with the highest perturbation budget. RPC-SymViT-base ($n$iter/all-layer) applies $n$ PAP iterations at all layers. }
    \label{tab:table10}
   \begin{tabular}{lcccc} 
    \toprule
      \multirow{2}{*}{Model} & \multicolumn{2}{c}{PGD} & \multicolumn{2}{c}{FGSM}\\
      & Top-1 $\uparrow$ & Top-5 $\uparrow$& Top-1 $\uparrow$ & Top-5 $\uparrow$ \\
        \midrule
      SymViT-base (baseline) & 12.11 & 25.55 & 53.61 & 78.05   \\
      \midrule
      RPC-SymViT-base (2iter/all-layer) & \textbf{13.41} & \textbf{26.68} & \textbf{54.68} & \textbf{78.06}
       \\
    \bottomrule
    \end{tabular}
  \end{center}
}
\end{table*}

\subsection{Results on RPC-Attention in FAN}
To validate that our RPC-Attention is also complementary with SOTA transformer models and can be combined with those methods to improve the SOTA results, we have conducted additional experiments in which we incorporate our RPC-Attention with FAN \cite{zhou2022understanding}. FAN is one of the top transformer models that obtain SOTA results on accuracy and robustness. A FAN model augments the MLP layer that follows the standard self-attention with a new channel attention (CA) block. This CA computes an attention matrix along the channel dimension, taking advantage of the feature covariance and allowing the model to filter out irrelevant information. 

We use the FAN-ViT-tiny (FAN-tiny) variant with a symmetric attention for training. In our RPC-Attention + FAN (RPC-FAN-tiny), we replace the attention blocks in the first layer of FAN with our RPC-Attention that runs 4 PAP iterations with hyperparameter values of $\lambda = 4$ and $\mu = ND /4\|\mK\|_1$. Both our RPC-FAN-tiny and the baseline FAN-tiny are trained for 300 epochs on the ImageNet-1K object classification task. We summarize our results in Tables \ref{tab:table11} and \ref{tab:table12}. RPC-FAN-tiny outperforms the baseline FAN-tiny on all evaluated benchmarks, including ImageNet-1k, ImageNet-R, and ImageNet-A. Additionally, on PGD and FGSM attacked data, RPC-FAN-tiny significantly outperforms FAN-tiny by over 3\%. 
\begin{table*}[!t]
{\footnotesize	
  \begin{center}
    \caption{\small Top-1, top-5 accuracy (\%) of RPC-FAN-tiny and FAN-tiny on clean ImageNet-1K data and popular standard robustness benchmarks for image classification. RPC-FAN-tiny ($n$iter/layer1) applies $n$ PAP iterations only at the first layer. } 
    \label{tab:table11}
    \begin{tabular}{lcccccc} 
    \toprule
      \multirow{2}{*}{Model} & \multicolumn{2}{c}{IN-1K} & IN-R & IN-A \\
      & Top-1 $\uparrow$ & Top-5 $\uparrow$ & Top-1 $\uparrow$ & Top-1 $\uparrow$ \\
        \midrule
      FAN-tiny (baseline) & 77.89 & 94.20 &  41.79& 13.40 \\
      \midrule
      RPC-FAN-tiny (4iter/layer1) & \textbf{77.98} & \textbf{94.27} & \textbf{42.02} & \textbf{13.55} \\
    \bottomrule
    \end{tabular}
  \end{center}
}
\end{table*}
\begin{table*}[t!]
{\footnotesize	
  \begin{center}
    \caption{\small Top-1 and top-5 accuracy (\%) of RPC-FAN-tiny and FAN-tiny on PGD and FGSM attacked ImageNet-1K validation data with the highest perturbation budget. RPC-FAN-ViT ($n$iter/layer1) applies $n$ PAP iterations only at the first layer. }
    \label{tab:table12}
    \begin{tabular}{lcccc} 
    \toprule
      \multirow{2}{*}{Model} & \multicolumn{2}{c}{PGD} & \multicolumn{2}{c}{FGSM}\\
      & Top-1 $\uparrow$ & Top-5 $\uparrow$& Top-1 $\uparrow$ & Top-5 $\uparrow$ \\
        \midrule
      FAN-tiny (baseline) & 2.92  & 4.86 & 32.01 & 61.72  \\
      \midrule
      RPC-FAN-tiny (4iter/layer1) &  \textbf{6.25}  & \textbf{10.01}  &  \textbf{35.12} & \textbf{63.40} 
       \\
    \bottomrule
    \end{tabular}
  \end{center}}
\end{table*}

\subsection{Results on downstream Natural Language Understanding tasks }
\label{appx:nlu}

\begin{table*}[!t]
  \begin{center}
    \caption{\small Train and validation accuracy (\%) of RPC-SymSMoE, SymSMoE, RPC-SMoE (asymmetric) and SMoE (asymmetric) on SST-2 and IMDB downstream tasks when pre-trained on WikiText-103. RPC is only implemented during fine-tuning, not pre-training and runs for 2 iterations for all layers in each model. }
    \label{tab:finetune}
{\footnotesize	
    \begin{tabular}{lcccc} 
    \toprule
      \multirow{2}{*}{Model} & \multicolumn{2}{c}{SST-2} & \multicolumn{2}{c}{IMDB}\\
      & Train Acc $\uparrow$ & Valid Acc $\uparrow$ & Train Acc $\uparrow$ & Valid Acc $\uparrow$\\
      \midrule
        SymSMoE (baseline) & 63.12 & 69.20 & -& -\\
      RPC-SymSMoE & \textbf{74.48} & \textbf{76.27} & -&-\\
      \midrule
      SMoE (baseline) & 53.20 & 69.38 & 64.12 & 65.74  \\
      RPC-SMoE & \textbf{58.87} &  \textbf{70.63} & \textbf{74.36} & \textbf{73.96} \\
    \bottomrule
    \end{tabular}}
  \end{center}
\end{table*}

\begin{table*}[!t]
  \begin{center}
    \caption{\small Validation and test accuracy of RPC-Attention implemented in a pre-trained transformer language model (RPC-LM) during fine-tuning versus the baseline transformer language model (Baseline-LM). We fine-tune both models on the 5-class Stanford Sentiment Treebank (SST-5) task with 2 iterations for all layers in the RPC model. }
    \label{tab:sst5}
{\footnotesize	
    \begin{tabular}{lcc}
        \toprule
        Model & Valid Acc $\uparrow$ & Test Acc $\uparrow$ \\
        \midrule
        Baseline-LM & 46.51 & 49.23 \\
        RPC-LM & \textbf{48.68} & \textbf{50.36} \\
        \bottomrule
    \end{tabular}}
  \end{center}
\end{table*}

We conduct additional experiments on several downstream Natural Language Understanding tasks to illustrate the effectiveness of RPC-Attention during fine-tuning as well. The Stanford Sentiment Treebank v2 (SST-2) and IMDB Sentiment Analysis (IMDB) tasks are binary classifications where the goal is to determine if sentences have positive or negative sentiments. We use 2 baseline Sparse Mixture of Experts (SMoE) models, one with symmetric attention (SymSMoE) and one with asymmetric attention (SMoE), pre-trained on WikiText-103 without RPC, then fine-tuned with (RPC-SymSMoE/RPC-SMoE) and without RPC-Attention for comparison. Their results can be found in Table \ref{tab:finetune}. We observe that RPC-models outperform the baseline models significantly on these tasks. 

On the 5-class sentiment classification task, Stanford Sentiment Treebank (SST-5), we use a pre-trained transformer language model from the NAACL 2019 tutorial on ``Transfer Learning in Natural Language Processing'' for fine-tuning. Their publicly available code can be found at \url{https://github.com/huggingface/naacl_transfer_learning_tutorial}. The objective of the task is to determine if the sentences have negative, somewhat negative, neutral, somewhat positive, or positive sentiments. We implement RPC-Attention during fine-tuning only (RPC-LM) and compare the results with the baseline model (Baseline-LM) on SST-5 in Table \ref{tab:sst5}. As can be seen from the table, our RPC-Attention is applicable to pre-trained language models and performs significantly better than the baseline. Hence, RPC-Attention is highly effective in downstream natural language understanding tasks and versatile in its application.

\subsection{Computational Efficiency}
\label{appx:eff}
A possible limitation of introducing an iterative scheme into a deep model is a significant increase in computational overhead. We aim to alleviate that concern by providing the number of flops per sample, run time per sample, memory and number of parameters of each RPC-SymViT variant and the SymViT baseline during both training and test time in Table \ref{tab:table13}. We observe that RPC-Attention is comparable to the baseline softmax attention across all metrics at test time while only slightly less efficient than the baseline in terms of the number of flops, run time per sample, and memory during training. 
\begin{table*}[t!]
{\footnotesize	
  \begin{center}
    \caption{\small Flop per sample, run time per sample, memory and number of parameters of each RPC-SymViT variant compared to the SymViT baseline. RPC-SymViT ($n$iter/layer1) applies $n$ PAP iterations only at the first layer. RPC-SymViT ($n$iter/all-layer) applies $n$ PAP iterations at all layers. }
    \label{tab:table13}
    \begin{tabular}{lccccccc}
\toprule
\multirow{2}{*}{Model} & \multirow{2}{*}{Flop/Sample} & Sec/Sample & Sec/Sample & Memory & Memory & \multirow{2}{*}{Parameters} \\
& & (Training) & (Test) & (Training) & (Test) & \\
\midrule
SymViT (baseline) & 1.17M & 0.079 & 0.010 & 1435MB & 1181MB & 5.2M \\
RPC-SymViT (4iter/layer1) & 1.22M & 0.082 & 0.010 & 1441MB & 1181MB & 5.2M \\
RPC-SymViT (5iter/layer1) & 1.23M & 0.084 & 0.010 & 1443MB & 1181MB & 5.2M \\
RPC-SymViT (6iter/layer1) & 1.25M & 0.085 & 0.011 & 1443MB & 1181MB & 5.2M \\
RPC-SymViT (2iter/layer) & 1.35M & 0.092 & 0.017 & 1461MB & 1181MB & 5.2M \\
\bottomrule
\end{tabular}
  \end{center}}
\end{table*}

\subsection{Results on Robust Asymmetric Attention}
\label{appx:rpc-asym}
\begin{table*}[t!]
{\footnotesize	
  \begin{center}
    \caption{\small Top-1, top-5 accuracy (\%) and AUPR of an implementation of RPC-Attention on asymmetric attention evaluated on ImagetNet-1K validation set, ImageNet-R, ImageNet-A and ImageNet-O. These results are compared to the standard asymmetric ViT. }
    \label{tab:table7}
    \begin{tabular}{lcccccc} 
    \toprule
      \multirow{2}{*}{Model} & \multicolumn{2}{c}{IN-1K} & IN-R & IN-A & IN-O \\
      & Top-1 $\uparrow$ & Top-5 $\uparrow$ & Top-1 $\uparrow$ & Top-1 $\uparrow$  & AUPR $\uparrow$\\
        \hline
      ViT & 72.11 & 90.97 & 32.41 & 7.65 & 17.36 \\
      \hline
      RPC-AsymViT (4iter/layer1) & 72.34 & 91.12 & 32.23 & 7.75 & 17.54  
       \\
       RPC-AsymViT (Sym/Asym) & 72.34 & 91.38 & 32.79 & 7.23 & 17.58 \\
    \bottomrule
    \end{tabular}
  \end{center}}
\end{table*}

In this section, we report the results of extending the RPC-SymViT model to the asymmetric attention. However, as the PAP Algorithm.~\ref{algo:pap} is not designed for multiple data matrices, it is not as effective in the asymmetric case. We implement two variations of the algorithm in an asymmetric attention ViT-tiny with 12 layers. In the 4iter/layer1 version, similar to the symmetric attention case, we run 4 iterations of the algorithm only in the 1st layer of the model by replacing $\text{Softmax}(\mK - \mS_{k+1} - \mu^{-1}\mY_k, \mK - \mS_{k+1} - \mu^{-1}\mY_k)$ with $\text{Softmax}(\mQ, \mK - \mS_{k+1} - \mu^{-1}\mY_k)$. For the second version, labeled Sym/Asym, we run the algorithm for 4 iterations in a symmetric attention mechanism in the 1st layer, followed by the standard asymmetric attention in layers 2 to 12. We compare these RPC-AsymViT models to the asymmetric softmax attention ViT. 

As we can see from Table~\ref{tab:table7}, both variants only improve slightly over the benchmark on most of the corrupted datasets. Such a result confirms our intuition that the ADMM algorithm does not extend easily to multiple corrupted matrices as it only solves an objective function involving a single low-rank data matrix and its sparse corruption matrix. 

\section{Further Discussion on Related Works}
\label{appx:related-works}
\cite{chen2023primalattention} provides a new perspective by emphasizing the asymmetry of the softmax kernel and recovers the self-attention mechanism from an asymmetric Kernel Singular Value Decomposition (KSVD) using the duality of the optimization problem. Separately, our kernel PCA perspective derives softmax attention and addresses the asymmetry of attention through a projection of the query vectors in feature space. While there are similarities between KSVD and kernel PCA, in the sense of finding low rank approximations and low dimensional representations of the data, the primal objective functions are different and we do not need to consider the dual form.
Another related work views transformers from the perspective of Support Vector Machines \cite{tarzanagh2023transformers}. Though, kernel PCA can be formulated in a similar fashion as a least squares SVM problem as explained in \cite{1189643}, our work focuses on the forward pass of attention and show that it can recover a kernel PCA solution. In contrast, \cite{1189643} examines the backward pass and optimization geometry of an attention layer towards a hard margin SVM solution that separates optimal tokens from non-optimal ones, under certain assumptions of the loss function, initial conditions, and certain SVM constraints. Furthermore, using our framework, we are able to predict the exact explicit form of the value matrix in self-attention, demonstrating that this matrix captures the eigenvectors of the Gram matrix of the key vectors in a feature space. To the best of our knowledge, our work is the first to show this insight. 

Hamburger in \cite{geng2021attention} models the global context discovery as the low-rank recovery of the input tensor and solves it via matrix decomposition. Both Hamburger and our Attention with Robust Principal Components (RPC-Attention) try to recover clean signal subspaces via computing a low-rank approximation of a given matrix. The key differences between our RPC-Attention and Hamburger are: (1) Our RPC-Attention finds a low-rank approximation of the key matrix while Hamburger finds a low-rank approximation of the input matrix, and (2) Our RPC-Attention models the corruption by a sparse matrix while Hamburger does not enforce this condition. The entries of this sparse corruption can have an arbitrarily large magnitude and help model grossly corrupted observations in which only a portion of the observation vector is contaminated by gross error. Numerous critical applications exist where the data being examined can be naturally represented as a combination of a low-rank matrix and a sparse contribution, such as video surveillance, face recognition, and collaborative filtering \cite{candes2009robust}. 

\cite{yang2021graph} derives each component in a Graph Neural Network (GNN) from the unfolded iterations of robust descent algorithms applied to minimizing a principled graph regularized energy function. In particular, propagation layers and nonlinear activations implement proximal gradient updates, and graph attention results from iterative reweighted least squares (IRLS). While this is an interesting approach, it has not been extended to explaining the architecture of transformers, including self-attention, yet. In contrast, our kernel principal component analysis (kernel PCA) allows us to derive self-attention in transformers, showing that the attention outputs are projections of the query vectors onto the principal components axes of the key matrix in a feature space. 

\cite{amos2017optnet} and \cite{bai2019deep} implement each layer as an optimization and fixed-point solver, respectively. In particular, an OptNet layer in \cite{amos2017optnet} solves a quadratic program, and a Deep Equilibrium layer in \cite{bai2019deep} computes the fixed point of a nonlinear transformation. Different from these layers, our RPC-Attention solves a Principal Component Pursuit - a convex program. Also, both OptNet layer in \cite{amos2017optnet} and Deep Equilibrium layer in \cite{bai2019deep} do not shed light on the derivation and formulation of self-attention, which our kernel PCA framework does. 

\section{Broader Impacts}\label{appendix: broader impacts}
Our research improves both clean data processing and robust performance, especially in areas with significant social relevance. Specifically, we demonstrate enhanced results in image segmentation, benefiting self-driving cars, and language modeling, enhancing AI chatbot assistants. We show notable improvements in resisting adversarial attacks, aiming to protect crucial AI systems from malicious actors. Additionally, we achieve competitive performance in language modeling with contaminated data, which is often encountered in real-world situations. Despite the potential for AI misuse, our research presents substantial advancements in fundamental architectures and theory, which we hope will inspire further socially beneficial developments.


\end{document}